%% file: newMain_2.tex
\documentclass[twoside,11pt]{article}

\usepackage[preprint]{jmlr2e}

\usepackage{svg}
\newcommand{\orcid}[1]{\href{https://orcid.org/#1}{\includesvg[width=10pt]{orcid}}}
\usepackage{amsmath,amssymb,amsfonts,mathrsfs}
\DeclareMathOperator*{\x}{{\boldsymbol{\theta}}}
\usepackage[linesnumbered,ruled,vlined]{algorithm2e}
\SetKwInput{KwInput}{Input}                
\SetKwInput{KwOutput}{Output}
\SetKwFor{when}{when}{do}{}

\usepackage{mathtools}

\DeclarePairedDelimiter\floor{\lfloor}{\rfloor}
\usepackage{color}
\usepackage{cancel}
\usepackage{afterpage}

\ShortHeadings{Incremental Hessian eigenvector sharing algorithm for federated learning}{Dal Fabbro, Dey, Rossi and Schenato}
\firstpageno{1}

\begin{document}
\title{SHED: A Newton-type algorithm for federated learning based on incremental Hessian eigenvector sharing}

\author{\name Nicol\`o Dal Fabbro \email nicolo.dalfabbro@studenti.unipd.it \\
       \addr Department of Information Engineering\\
       University of Padova\\
       Padova, PD 35122, Italy
       \AND
      \name Subhrakanti Dey \email  subhrakanti.dey@angstrom.uu.se \\
      \addr Department of Electrical Engineering\\
      Uppsala University\\
      Uppsala, 751 03, Sweden
      \AND
      \name Michele Rossi \email michele.rossi@unipd.it\\
      \addr Department of Information Engineering\\
       University of Padova\\
       Padova, PD 35122, Italy
       \AND
       \name Luca Schenato \email schenato@dei.unipd.it\\  
       \addr Department of Information Engineering\\
       University of Padova\\
       Padova, PD 35122, Italy
      }

\editor{}

\maketitle

\begin{abstract}
There is a growing interest in the distributed optimization framework that goes under the name of Federated Learning (FL). In particular, much attention is being turned to FL scenarios where the network is strongly heterogeneous in terms of communication resources (e.g., bandwidth) and data distribution. In these cases, communication between local machines (\emph{agents}) and the central server (\emph{Master}) is a main consideration. In this work, we present SHED\footnote{In a previous preprint version of this paper, the algorithm was referred to as IOS}, an original communication-constrained Newton-type (NT) algorithm designed to accelerate FL in such heterogeneous scenarios. SHED is by design robust to non i.i.d. data distributions, handles heterogeneity of agents' communication resources (CRs), only requires sporadic Hessian computations, and achieves super-linear convergence. This is possible thanks to an \emph{incremental} strategy, based on eigendecomposition of the local Hessian matrices, which exploits (possibly) \emph{outdated} second-order information. The proposed solution is thoroughly validated on real datasets by assessing (i) the number of communication rounds required for convergence, (ii) the overall amount of data transmitted and (iii) the number of local Hessian computations. For all these metrics, the proposed approach shows superior performance against state-of-the art techniques like GIANT and FedNL.
\end{abstract}
\section{Introduction}
\label{sec:intro}
\input{JMLR/section_1}

\section{Problem formulation}
\label{sec:probForm}
\input{JMLR/section_2}

\section{Linear regression (least squares)}\label{sec:AlgsAndTheory}
\input{JMLR/section_3}

\section{From least squares to strongly convex cost}\label{sec:fromLStoConvex}
\input{JMLR/section_4}

\section{Federated learning with convex cost}\label{sec:generalConvex}
\input{JMLR/section_5}

\section{Empirical Results}\label{sec:results}
\input{JMLR/section_6}

\section{Conclusions}

In this work, we have proposed SHED, a Newton-type algorithm to perform FL in heterogeneous communication networks. SHED is versatile with respect to agents' (per-iteration) communication resources and operates effectively in the presence of {\it non i.i.d.} data distributions, outperforming state-of-the-art techniques. 
It achieves better performance with respect to the competing FedNL approach, while involving sporadic Hessian computations. In the case of i.i.d. data statistics, SHED is also competitive with GIANT, even though the latter may perform better under certain conditions. We stress that the key advantage of SHED lies in its robustness under any data distribution, and its effectiveness and versatility when communication resources differ across nodes and links.

Future work includes the use and extension of the algorithm for more specific scenarios and applications, like, for example, wireless networks, and also the study of new heuristics for the renewal operation in the general convex case. Furthermore, future research directions may deal with compression techniques for the Hessian eigenvectors, as well as the development of heuristic algorithms combining different second-order methods, like FedNL and SHED. Lastly, the spectral characteristics of the data could also be exploited to tailor the proposed algorithm to specific FL problems, leading to further benefits in terms of convergence speed.


\acks{This work has been supported, in part, by the Italian Ministry of Education, University and Research, through the PRIN project no. 2017NS9FEY entitled ``Realtime Control of 5G Wireless Networks: Taming the Complexity of Future Transmission and Computation Challenges''.}


\appendix
\section*{Appendix A.}\label{app:theorems}
\input{JMLR/Appendix_A}
\section*{Appendix B.}\label{app:theorems}
\input{JMLR/Appendix_B}
\afterpage{\null\newpage}
\newpage
\vskip 0.2in
\bibliography{biblio}

\end{document}

%% file: JMLR/section_1.tex
With the growing computational power of edge devices and the booming increase of data produced and collected by users worldwide, solving machine learning problems without having to collect data at a central server is becoming very appealing \citep{EdgeAI}. One of the main reasons for not transferring users' data to cloud central servers is due to privacy concerns. Indeed, users, such as individuals or companies, may not want to share their private data with other network entities, while training their machine learning algorithms. In addition to privacy, distributed processes are by nature more resilient to node/link failures and can be directly implemented on the servers at network edge, i.e., within multi-access edge computing (MEC) scenarios \citep{MEC-2020}.

This distributed learning framework goes under the name of federated learning (FL) (\citealp{McMahanFL, GENERAL_FL}), and it has attracted much research interest in recent years. Direct applications of FL can be found, for example, in the field of healthcare systems \citep{CARE_1, CARE_3} or of smartphone utilities \citep{KEYWORD_PRED}.

One of the most popular FL algorithms is federated averaging (FedAvg) \citep{McMahanFL}, which achieves good results, but which comes with weak convergence guarantees and has been shown to diverge under heterogeneous data distributions \citep{HeterogeneousDiv}.

Indeed, among the open challenges of FL, a key research question is how to provide efficient distributed optimization algorithms in scenarios with heterogeneous communication links (different bandwidth) and non i.i.d. data distributions \citep{advances, nonIidSurvey, MULTITASK_HETER, FL_ON_NONIID}. These aspects are found in a variety of applications, such as the so-called \emph{federated edge learning} framework, where learning is moved to the network edge, and which often involves unstable and heterogeneous wireless connections \citep{EdgeAI, DONE, AcceleratingComLet, JointWirelessFL}. In addition, the MEC and FL paradigms are of high interest to IoT scenarios such as data retrieval and processing within smart cities, which naturally entail non i.i.d. data distributions due to inherent statistical differences in the underlying spatial processes (e.g., vehicular mobility, user density, etc.) \citep{MEC-IoT-2020}. 

The bottleneck represented by communication overhead is one of the most critical aspects of FL. In fact, in scenarios with massive number of devices involved, inter-agent communication can be much slower than the local computations performed by the FL agents themselves (e.g., the edge devices), by many orders of magnitude \citep{GENERAL_FL}. The problem of reducing the communication overhead of FL becomes even more critical when the system is characterized by non i.i.d. data distributions and heterogeneous communication resources (CRs) \citep{HeterogeneousDiv}. In this setting, where the most critical aspect is inter-node communications, FL agents are usually assumed to have good computing capabilities, and wisely increasing the computation effort at the agents is a good strategy to obtain a faster convergence. For this reason, Newton-type approaches, characterized by strong robustness and fast convergence rates, even if computationally demanding, have been recently advocated \citep{LocNewton, FedNL}.

In the present work, we present SHED (Sharing Hessian Eigenvectors for distributed learning), a Newton-type FL framework that is naturally robust to non i.i.d. data distributions and that intelligently allocates the (per-iteration) communication resources of the involved FL agents, allowing those with more CRs to contribute more towards improving the convergence rate. Differently from prior art, SHED requires FL agents to locally compute the Hessian matrix sporadically. Its super-linear convergence is here proven by studying the CRs-dependent convergence rate of the algorithm by analyzing the dominant Lypaunov exponent of the estimation error. With respect to other Newton-type methods for distributed learning, our empirical results demonstrate that the proposed framework is (i) competitive with state-of-the-art approaches in i.i.d. scenarios and (ii) robust to non i.i.d. data distributions, for which it outperforms competing solutions.

\subsection{Related work}
Next, to put our contribution into context, we review previous related works on first and second order methods for distributed learning.

\textbf{First-order methods.} Some methods have been recently proposed to tackle non i.i.d. data and heterogeneous networks. SCAFFOLD \citep{scaffold}, FedProx \citep{HeterogeneousDiv} and the work in \cite{SGD_VAR_REstart} propose modifications to the FedAvg algorithm to face system's heterogeneity. To deal with non i.i.d. datasets, strategies like those in \cite{FL_ON_NONIID} and \cite{jeong2018} have been put forward, although these approaches require that some data is shared with the central server, which does not fit the privacy requirements of FL. To reduce the FL communication overhead, the work in \cite{LAG} leverages the use of outdated first-order information by designing simple rules to detect slowly varying local gradients. With respect to heterogeneous time-varying CRs problem, \cite{Gunduz} presented techniques based on gradient quantization and on analog communication strategies exploiting the additive nature of the wireless channel, while \cite{JointWirelessFL} studied a framework to jointly optimize learning and communication for FL in wireless networks.

\textbf{Second-order methods.} Newton-type (NT) methods exploit second-order information of the cost function to provide accelerated optimization, and are therefore appealing candidates to speed up FL. NT methods have been widely investigated for distributed learning purposes: GIANT \citep{GIANT} is an NT approach exploiting the harmonic mean of local Hessian matrices in distributed settings. Other related techniques are LocalNewton \citep{LocNewton}, DANE \citep{DANE}, AIDE \citep{AIDE}, DiSCO \citep{DISCO}, DINGO \citep{DINGO} and DANLA \citep{DANLA}. DONE \citep{DONE} is another technique inspired by GIANT and specifically designed to tackle federated edge learning scenarios. Communication efficient NT methods like GIANT and DONE exploit an extra communication round to obtain estimates of the global Hessian from the harmonic mean of local Hessian matrices. These algorithms, however, were all designed assuming that data is i.i.d. distributed across agents and, as we empirically show in this paper, under-perform if such assumption does not hold. A recent study, FedNL \citep{FedNL}, proposed algorithms based on matrix compression which use theory developed in \cite{Islamov} to perform distributed training, by iteratively learning the Hessian matrix at the optimum. A similar approach based on matrix compression techniques has been considered in Basis Learn \citep{BasisLearn}, proposing variants of FedNL that exploit change of basis in the space of matrices to improve the quality of the compressed Hessian. However, these solutions do not consider heterogeneity in the CRs, and require the computation of the local Hessian at each iteration. Another NT technique has been recently proposed in \cite{FEEL}, based on the popular L-BFGS \citep{L-BFGS} quasi-Newton method. Other recent related NT approaches have been proposed in FLECS \citep{FLECS}, FedNew \citep{FedNew} and Quantized Newton \citep{QuantizedNewt}. Although some preliminary NT approaches have been proposed for the FL framework, there is still large space for improvements especially in ill-conditioned setups with heterogeneous CRs and non i.i.d. data distributions.

\subsection{Contribution}

In this work, we design SHED (Sharing Hessian Eigenvectors for Distributed learning), a communication-constrained NT optimization algorithm for FL that is suitable for networks with heterogeneous CRs and non i.i.d. data. SHED requires only sporadic Hessian computation and it is shown to accelerate convergence of convex federated learning problems, reducing communication rounds and communication load in networks with heterogeneous CRs and non i.i.d. data distributions.
    
 In SHED, local machines, the \emph{agents}, compute the local Hessian and its eigendecomposition. Then, according to their available communication resources (CRs), agents share some of the most \emph{relevant} Hessian eigenvector-eigenvalue pairs (EEPs) with the server (the Master). Together with the EEPs, agents send to the Master a scalar quantity - computed locally - that allows a full-rank Hessian approximation. The obtained approximations are averaged at the Master in order to get an approximated global Hessian, to be used to perform global NT descent steps. The approximation of the Hessian matrix at the Master is then improved by \emph{incrementally} sending additional EEPs. In particular, for a general convex problem, SHED is designed to make use of \emph{outdated} Hessian approximations. Thanks to this, local Hessian computation is required only sporadically. We analyze the convergence rate via the analysis of the dominant Lyapunov exponent of the objective parameter estimation error. In this novel framework, we show that the use of outdated Hessian approximations provides an improvement in the convergence rate. Furthermore, we show that the EEPs of the outdated local Hessian approximations can be shared heterogeneously with the Master, so the algorithm is versatile with respect to (per-iteration) heterogeneous CRs. We also show that, thanks to the proposed incremental strategy, SHED enjoys a super-linear convergence rate.


\subsection{Organization of the paper}

The rest of the paper is organized as follows: in Section \ref{sec:probForm}, we detail the problem formulation and the general idea behind SHED design. In Section \ref{sec:AlgsAndTheory}, we present SHED and the theoretical results in the linear regression (least squares) case, describing the convergence rate with respect to the algorithm parameters. Section \ref{sec:fromLStoConvex} is instrumental to extending the algorithm to a general strongly convex problem, which is done later on in Section \ref{sec:generalConvex}. At the end of Section \ref{sec:generalConvex}, based on the theoretical results, we propose some heuristic choices for tuning SHED parameters. Finally, in Section \ref{sec:results}, empirical performance of SHED are shown on real datasets. In particular, the theoretical results are validated and compared against those from state-of-the art approaches. Longer proofs and some additional experiments are reported in the Appendices, for completeness.

\emph{Notation:} Vectors and matrices are written as lower and upper case bold letters, respectively (e.g., vector $\mathbf{v}$ and matrix $\mathbf{V}$). The operator $\|\cdot\|$ denotes the $2$-norm for vectors and the spectral norm for matrices. $diag(\mathbf{v})$ denotes a diagonal matrix with the components of vector $\mathbf{v}$ as diagonal entries. $\mathbf{I}$ denotes the identity matrix.

%% file: JMLR/section_2.tex
\begin{figure}[t]
\centering
		\includegraphics[width=0.65\columnwidth]{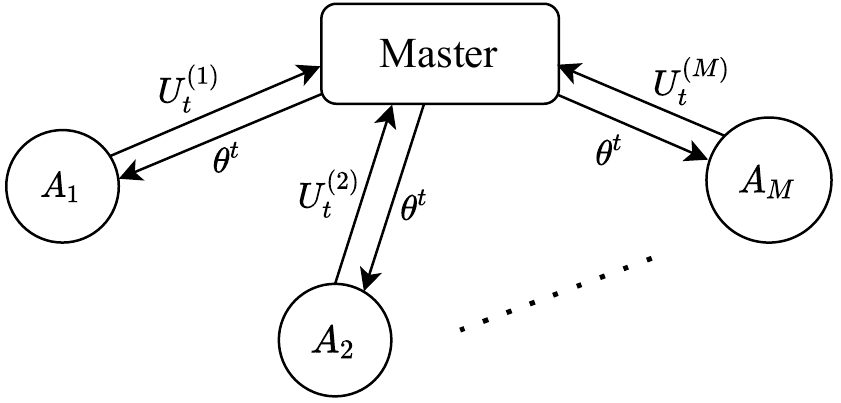}
		\caption{The typical FL optimization framework: agents $A_1, ..., A_M$ cooperate to solve a common learning problem. After receiving the global parameter $\x^t$, at the current iteration $t$, they share their optimization sets $U_t^{(1)}, ..., U_t^{(M)}$ so that a new global update can be performed at the Master.}
		\label{fig:frameworkFL}
\end{figure}

In Fig. \ref{fig:frameworkFL} we illustrate the typical framework of a federated learning scenario. In the rest of the paper, we denote the optimization parameter by $\boldsymbol{\theta}\in\mathbb{R}^n$, and a generic cost function by $f:\mathbb{R}^n \rightarrow \mathbb{R}$, where $n$ is the dimension of the feature data vectors $\mathbf{x} \in \mathbb{R}^n$.\newline
Let us denote a dataset by $\mathcal{D} = \{\mathbf{x}_j, y_j\}_{j = 1}^{N}$, where $N$ is the global number of data samples. $\mathbf{x}_j \in \mathbb{R}^n$ denotes the $j$-th data sample and $y_j \in \mathbb{R}$ the response of sample $j$. In the case of classification problems, $y_j$ would be an integer specifying the class to which sample $\mathbf{x}_j$ belongs.
We consider the problem of regularized empirical risk minimization of the form:
\begin{equation}\label{eq:sumOfConvexFunctions}
    \min_{\boldsymbol{\theta}}f(\boldsymbol{\theta}),
\end{equation}
with
\begin{equation*}
    f(\boldsymbol{\theta}) := \frac{1}{N}\sum_{j = 1}^{N}l_j(\boldsymbol{\theta}) + \frac{\mu}{2}\|\boldsymbol{\theta}\|_2^2,
\end{equation*}
where each $l_j(\boldsymbol{\theta})$ is a convex function related to the $i$-th element of $\mathcal{D}$ and $\mu$ a regularization parameter. In particular, 
\begin{equation*}
l_j(\boldsymbol{\theta}) = l(\mathbf{x}_j^T\boldsymbol{\theta}, y_j), 
\end{equation*}
where $l: \mathbb{R} \rightarrow\mathbb{R}$ is a convex function. Examples of convex cost functions considered in this work are:
\begin{itemize}
    \item linear regression with quadratic cost (least squares):
    \begin{equation*}\label{}
    l(\mathbf{x}_j^T\boldsymbol{\theta}, y_j) = \frac{1}{2}(\mathbf{x}_j^T\boldsymbol{\theta} - y_j)^2    
    \end{equation*}
    
    \item logistic regression:
    \begin{equation*}
        l(\mathbf{x}_j^T\boldsymbol{\theta}, y_j) = \log(1 + e^{-y_j(\mathbf{x}_j^T\boldsymbol{\theta})})
    \end{equation*}
\end{itemize}
Logistic regression belongs to the class of generalized linear models. In general, in this paper we consider optimization problems in which the following assumptions on the agents' cost functions $f^{(i)}$ hold:
\newtheorem{assumption}{Assumption}
\begin{assumption}\label{ass:localLCont}
Let $\mathbf{H}^{(i)}(\boldsymbol{\theta}) := \nabla^2f^{(i)}(\boldsymbol{\theta})$ the local Hessian matrix of the cost $f^{(i)}(\x)$ of agent $i$. $f^{(i)}(\x)$ is twice continuously differentiable, $K_i$-smooth, $\kappa_i$-strongly convex and $\mathbf{H}^{(i)}(\boldsymbol{\theta})$ is $L_i$-Lipschitz continuous for all $i = 1, ..., M$.
\end{assumption}
The above assumptions imply that
\begin{equation*}
     \kappa_i \mathbf{I}\leq \mathbf{H}^{(i)}(\boldsymbol{\theta})  \leq K_i\mathbf{I}, \ \forall{\boldsymbol{\theta}}
\end{equation*}
\begin{equation*}
    \|\mathbf{H}^{(i)}(\boldsymbol{\theta}) - \mathbf{H}^{(i)}(\boldsymbol{\theta}')\| \leq L_i\|\boldsymbol{\theta} - \boldsymbol{\theta}'\|, \ \forall{\boldsymbol{\theta}, \boldsymbol{\theta}'}
\end{equation*}
Note that $K_i$-smoothness is also a consequence of the $K$-smoothness of any global cost function $f$ for some constant $K$, while strong convexity of agents' cost functions is always guaranteed if the functions $l_j(\x)$ are convex in the presence of a regularization term $\mu > 0$.

In the paper, we denote the data matrix by $\mathbf{X} = [\mathbf{x}_1, ..., \mathbf{x}_N] \in \mathbb{R}^{n\times N}$, and the label vector as $\mathbf{Y} = [y_1, ..., y_N] \in \mathbb{R}^{1\times N}$. We refer to the dataset $\mathcal{D}$ as the tuple $\mathcal{D} = (\mathbf{X}, \mathbf{Y})$. We denote by $M$ the number of agents involved in the optimization algorithm, and we can write $\mathbf{X} = [\mathbf{X}_1, ..., \mathbf{X}_M]$, $\mathbf{Y} = [\mathbf{Y}_1, ..., \mathbf{Y}_M]$, where $\mathbf{X}_i \in \mathbb{R}^{n\times N_i}$ and $\mathbf{Y}_i \in \mathbb{R}^{1\times N_i}$, where $N_i$ is the number of data samples of the $i$-th agent. We denote the local dataset of agent $i$ by $\mathcal{D}_i = (\mathbf{X}_i, \mathbf{Y}_i)$.
\subsection{A Newton-type method based on eigendecomposition}
The Newton method to solve (\ref{eq:sumOfConvexFunctions}) works as follows (Newton update):
\begin{equation*}
    \boldsymbol{\theta}^{t+1} = \boldsymbol{\theta}^{t} - \eta_t\mathbf{H}_t^{-1}\mathbf{g}_t,
\end{equation*}
where $t$ denotes the $t$-th iteration, $\mathbf{g}_t = \mathbf{g}(\boldsymbol{\theta}_t) = \nabla f (\boldsymbol{\theta}^t)$ is the gradient at iteration $t$ and $\eta_t$ is the step size (also called learning rate) at iteration $t$. $\mathbf{H}_t = \nabla^2 f (\boldsymbol{\theta}^t)$ denotes the Hessian matrix at iteration $t$. Compared to gradient descent, the Newton method exploits the curvature information provided by the Hessian matrix to improve the descent direction. We define $\mathbf{p}_t := \mathbf{H}_t^{-1}\mathbf{g}_t$. In general, Newton-type (NT) methods try to get an approximation of $\mathbf{p}_t$. In an FL scenario, assuming for simplicity that all $M$ agents have the same amount of data, we have that:
\begin{equation}
    \mathbf{H}_t = \frac{1}{M}\sum_{i = 1}^{M}\mathbf{H}_t^{(i)}, \ \ \mathbf{g}_t = \frac{1}{M}\sum_{i = 1}^{M}\mathbf{g}_t^{(i)},
\end{equation}
where $\mathbf{H}_t^{(i)} = \nabla^2f^{(i)}(\boldsymbol{\theta}^t)$ and $\mathbf{g}_t^{(i)} = \nabla f^{(i)}(\boldsymbol{\theta}^t)$ denote local Hessian and gradient of the local cost $f^{(i)}(\boldsymbol{\theta}^t)$ of agent $i$, respectively. 
To get a Newton update at the master, in an FL setting one would need each agent to transfer the whole matrix $\mathbf{H}_t^{(i)}$ of size  $O(n^2)$ to the master at each iteration, that is considered a prohibitive communication complexity in a federated learning setting, especially when the size of the feature data vectors, $n$, increases. Because of this, in this work we propose an algorithm in which the Newton-type update takes the form:
\begin{equation}\label{eq:qN-update}
    \boldsymbol{\theta}^{t+1} = \boldsymbol{\theta}^t - \eta_t\hat{\mathbf{H}}_t^{-1}\mathbf{g}_t,
\end{equation}
where $\hat{\mathbf{H}}_t$ is an approximation of $\mathbf{H}_t$. In some previous contributions, like \cite{GIANT, DONE}, $\hat{\mathbf{H}}_t$ is the harmonic mean of local Hessians, obtained at the master through an intermediate additional communication round to provide each agent with the global gradient $\mathbf{g}_t$. In the recent FedNL algorithm \citep{FedNL}, each agent at each iteration sends a compressed version of a Hessian-related matrix. In the best performing version of the algorithm, the rank-1 approximation using the first eigenvector of the matrix is sent. Authors show that thanks to this procedure they can eventually get the Hessian matrix at the optimum, which provides super-linear convergence rate.

Instead, our proposal is to incrementally obtain at the master an average of full-rank approximations of local Hessians through the communication of the local Hessian most relevant eigenvectors together with a carefully computed local approximation parameter. In particular, we approximate the Hessian matrix $\mathbf{H}_t$ exploiting eigendecomposition in the following way: the symmetric positive definite Hessian $\mathbf{H}_t$ can be diagonalized as $\mathbf{H}_t = \mathbf{V}_t\boldsymbol{\Lambda}_t \mathbf{V}_t^T$, with $\boldsymbol{\Lambda}_t = diag(\lambda_{1, t}, ..., \lambda_{n, t})$, where $\lambda_{k, t}$ is the eigenvalue corresponding to the $k$-th eigenvector,  $\mathbf{v}_{k, t}$. $\mathbf{H}_t$ can be approximated as
\begin{equation}\label{eq:FullApprox}
    \hat{\mathbf{H}}_t := \hat{\mathbf{H}}_t(\rho_t, q_t) =  \mathbf{V}_t\hat{\boldsymbol{\Lambda}}_t\mathbf{V}_t^T = \sum_{k = 1}^{q_t}(\lambda_{k, t} - \rho_t)\mathbf{v}_{k, t}\mathbf{v}_{k, t}^T + \rho_t \mathbf{I},
\end{equation}
with $\hat{\boldsymbol{\Lambda}}_t := \hat{\boldsymbol{\Lambda}}_t(\rho_t, q_t) = diag(\lambda_{1, t}, ..., \lambda_{q_t}, \rho_t, ..., \rho_t)$. The scalar $\rho_t > 0$ is the approximation parameter (if $\rho_t = 0$ this becomes a low-rank approximation). The integer $q_t = 1, ..., n$ denotes the number of eigenvalue-eigenvector pairs (EEPs) $\{\lambda_{k, t}, \mathbf{v}_{k, t}\}_{k =1}^{q_t}$ being used to approximate the Hessian matrix. We always consider eigenvalues ordered so that $\lambda_{1, t} \geq \lambda_{2, t}\geq  ... \geq \lambda_{n, t}$. The approximation of Eqn. (\ref{eq:FullApprox}) was used in \cite{MontanariWay} for a subsampled centralized optimization problem. There, the parameter $\rho_t$ was chosen to be equal to $\lambda_{q_t+1}$. In the FL setting, we use approximation shown in Eqn. (\ref{eq:FullApprox}) to approximate the local Hessian matrices of the agents. In particular, letting $\hat{\mathbf{H}}_t^{(i)}$ be the approximated local Hessian of agent $i$, the approximated global Hessian is the average of the local approximated Hessian matrices:
\begin{equation*}
    \hat{\mathbf{H}}_t = \sum_{i=1}^M{p_i\hat{\mathbf{H}}_t^{(i)}}, \ \ p_i = N_i/N
\end{equation*}
where $\hat{\mathbf{H}}_t^{(i)} := \hat{\mathbf{H}}_t(\rho_t^{(i)}, q_t^{(i)})$ is a function of the local approximation parameter $\rho_t^{(i)}$ and of the number of eigenvalue-eigenvector pairs $q_t^{(i)}$ shared by agent $i$, which are denoted by $\{\lambda_{k, t}^{(i)}, \mathbf{v}_{k, t}^{(i)}\}_{k =1}^{q_t^{(i)}}$.
\subsection{The algorithm in a nutshell}
The idea of the SHED algorithm is that agents share with the Master, together with the gradient, some of their local Hessian EEPs, according to the available CRs. They share the EEPs in a decreasing order dictated by the value of the positive eigenvalues corresponding to the eigenvectors. At each iteration, they incrementally add new EEPs to the information they have sent to the Master. In a linear regression problem, in which the Hessian does not depend on the current parameter, agents would share their EEPs incrementally up to the $n$-th. When the $n$-th EEP is shared, the Master has the full Hessian available and no further second order information needs to be transmitted. In a general convex problem, in which the Hessian matrix changes at each iteration, as it is a function of the current parameter, SHED is designed in a way in which agents perform a \emph{renewal} operation at certain iterations, i.e., they re-compute the Hessian matrix and re-start sharing the EEPs from the most relevant ones of the new matrix.

%% file: JMLR/section_3.tex

In this section we illustrate our algorithm and present the convergence analysis considering the problem of solving (\ref{eq:sumOfConvexFunctions}) via Newton-type updates (\ref{eq:qN-update}) in the least squares (LSs) case, i.e., in the case of linear regression with quadratic cost. Before moving to the FL case, we prove some results for the convergence rate of the centralized iterative least squares problem. \newline First, we provide some important definitions.
In the case of linear regression with quadratic cost, the Hessian is such that $\mathbf{H}_{LS} := \mathbf{H}(\boldsymbol{\theta}), \  \forall{\boldsymbol{\theta}}$, i.e., the Hessian does not depend on the parameter $\boldsymbol{\theta}$. Because of this, when considering LSs, we write the eigendecomposition as $\mathbf{H}_{LS} = \mathbf{V}\boldsymbol{\Lambda}\mathbf{V}^T$, with $\boldsymbol{\Lambda} = diag(\lambda_{1}, ..., \lambda_{n})$ without specifying the iteration $t$ when not needed. Let $\boldsymbol{\theta}^*$ denote the solution to (\ref{eq:sumOfConvexFunctions}). In the following, we use the fact that the cost for parameter $\boldsymbol{\theta}^t$ can be written as $f(\boldsymbol{\theta}^t) = f(\boldsymbol{\theta}^*) +  \bar{f}(\boldsymbol{\theta}^t)$, with $\bar{f}(\boldsymbol{\theta}^t) := \frac{1}{2}(\boldsymbol{\theta}^t - \boldsymbol{\theta}^*)^T\mathbf{H}_{LS}(\boldsymbol{\theta}^t - \boldsymbol{\theta}^*)$. Similarly, the gradient can be written as $\mathbf{g}_t = \mathbf{H}_{LS}(\boldsymbol{\theta}^t - \boldsymbol{\theta}^*)$.\newline
In this setup, the update rule of Eqn. (\ref{eq:qN-update}) can be written as a time-varying linear discrete-time system:
\begin{equation}\label{eq:DT_system}
    {\x}^{t+1} - {\x}^* = \mathbf{A}_t({\x}^t- {\x}^*)
\end{equation}
where 
\begin{equation*}
    \mathbf{A}_t := \mathbf{A}(\rho_t, \eta_t, q_t) = \mathbf{I} - \eta_t\hat{\mathbf{H}}_t^{-1}\mathbf{H}_{LS}.
\end{equation*}
Indeed,
\begin{equation*}
\begin{aligned}
    {\x}^{t+1} - {\x}^* &= {\x}^{t} - {\x}^* - \eta_t\hat{\mathbf{H}}_t^{-1}\mathbf{g}_t \\&= \boldsymbol{\theta}^{t} - \boldsymbol{\theta}^* - \eta_t\hat{\mathbf{H}}_t^{-1}\mathbf{H}_{LS}(\boldsymbol{\theta}^t - \boldsymbol{\theta}^*)\\& = (\mathbf{I} - \eta_t\hat{\mathbf{H}}_t^{-1}\mathbf{H}_{LS})(\boldsymbol{\theta}^t - \boldsymbol{\theta}^*).
\end{aligned}
\end{equation*}
\subsection{Centralized iterative least squares}\label{sec:centralLSs}
In this sub-section, we study the optimization problem in the centralized case, so when all the data is kept in a single machine. We provide a range of choices for the approximation parameter $\rho_t$ that are optimal in the convergence rate sense. We denote the convergence factor of the descent algorithm described by Eqn. (\ref{eq:qN-update}) by
\begin{equation*}
r_t := r(\rho_t, \eta_t, q_t),    
\end{equation*}
making its dependence on the tuple $(\rho_t, \eta_t, q_t)$ explicit.
\begin{theorem}\label{lemma_1}
Consider solving problem (\ref{eq:sumOfConvexFunctions}) via Newton-type updates (\ref{eq:qN-update}) in the least squares case. At iteration $t$, let the Hessian matrix $\mathbf{H}_{LS}$ be approximated as in Eqn. (\ref{eq:FullApprox}) (centralized case).
The convergence rate is described by
\begin{equation}\label{tightBound}
    \|\boldsymbol{\theta}^{t+1} - \boldsymbol{\theta}^*\| \leq r_t\|\boldsymbol{\theta}^{t} - \boldsymbol{\theta}^*\|.
\end{equation}
For a given $q_t \in \{0, 1, ..., n\}$ the best achievable convergence factor is
\begin{equation}\label{eq:bestRate}
    r_t^* = r^*(q_t):= \min_{(\rho_t, \eta_t)}{r(\rho_t, \eta_t, q_t)} = (1 - \frac{\lambda_n}{\rho^*_t}),
\end{equation}
where
\begin{equation}\label{eq:defRhoStar}
    \rho_t^* := (\lambda_{q_t+1} + \lambda_{n})/2.
\end{equation}
$r_t^*$ is achievable if and only if $(\rho_t, \eta_t) \in \mathcal{S}^*$, with
\begin{equation}
    \mathcal{S}^* = \{(\rho_t, \eta_t): \rho_t \in [\lambda_n, \lambda_{q_t + 1}], \eta_t^* = \frac{2\rho_t}{\lambda_{q_t+1} + \lambda_n}\}
\end{equation}
\begin{proof}
From (\ref{eq:FullApprox}), writing $\mathbf{H}_{LS} = \mathbf{V}\boldsymbol{\Lambda}\mathbf{V}^T$ and $\hat{\mathbf{H}}_{t} = \mathbf{V}\hat{\boldsymbol{\Lambda}}_{t}\mathbf{V}^T$, with $\hat{\boldsymbol{\Lambda}}_{t} = diag(\lambda_1, ..., \lambda_{q_t}, \rho_t, ..., \rho_t)$, define $\boldsymbol{\theta}_{\rho_t, \eta_t}^{t+1} := \boldsymbol{\theta}^t - \eta_t{\hat{\mathbf{H}}_{t}}^{-1}\mathbf{g}_t$. Recalling that $\mathbf{g}_t = \mathbf{H}_{LS}(\boldsymbol{\theta}^t - \boldsymbol{\theta}^*)$, we have:
\begin{equation}\label{eq:defAlgUpdate}
\begin{aligned}
\boldsymbol{\theta}_{\rho_t, \eta_t}^{t+1} - \boldsymbol{\theta}^* =  \mathbf{A}_t(\boldsymbol{\theta}^t - \boldsymbol{\theta}^*) &=  (\mathbf{I} - \eta_t\hat{\mathbf{H}}_{t}^{-1}\mathbf{H}_{LS})(\boldsymbol{\theta}^t - \boldsymbol{\theta}^*)\\&=
\mathbf{V}(\mathbf{I} - \eta_t\hat{\boldsymbol{\Lambda}}_{t}^{-1}\boldsymbol{\Lambda})\mathbf{V}^T(\boldsymbol{\theta}^t - \boldsymbol{\theta}^*).
\end{aligned}
\end{equation}

For some given $q_t \in \{1, ..., n\}$, $r_t$ is a function of two tunable parameters, i.e., the tuple $(\eta_t, \rho_t)$.
We now prove that $r_t^*$ can be achieved if and only if $\rho_t \in [\lambda_n, \lambda_{q_t+1}]$. The convergence rate is determined by the eigenvalue of $(\mathbf{I} - \eta_t\hat{\boldsymbol{\Lambda}}_{t}^{-1}\boldsymbol{\Lambda})$ with the greatest absolute value. First, we show that $\rho_t \notin [\lambda_{n}, \lambda_{q_t + 1}]$ implies $r_t > r_t^*$, then we show that, if $\rho_t \in [\lambda_{n}, \lambda_{q_t + 1}]$, there exists an optimal $\eta_t^*$ for which $r_t^*$ is achieved. If $\rho_t < \lambda_{n}$, the choice of $\eta_t$ minimizing the maximum absolute value of $(\mathbf{I} - \eta_t\hat{\boldsymbol{\Lambda}}_{t}^{-1}\boldsymbol{\Lambda})$ is the solution of $|1 - \eta_t| = |1 - \eta_t\lambda_{q_{t+1}}/\rho_t|$, which is $\eta_t^* = 2\rho_t/(\rho_t + \lambda_{q_t+1})$. The corresponding convergence factor is $1 - \eta_t^* > r_t^*$. Similarly, if $\rho_t > \lambda_{q_t+1}$, one gets $\eta_t^* = 2\rho_t/(\rho_t + \lambda_n)$ and convergence factor equal to $1 - 2\lambda_n/(\rho_t + \lambda_n) > r_t^*$. If $\rho_t \in [\lambda_n, \lambda_{q_t+1}]$, the best $\eta_t$ is such that $|1 - \eta_t\lambda_{n}/\rho_t| = |1 - \eta_t\lambda_{q_{t+1}}/\rho_t|$, whose solution is
\begin{equation}\label{eq:bestEta}
    \eta_t^* = \frac{2\rho_t}{\lambda_{q_t+1} + \lambda_n}
\end{equation}
and the achieved factor is $r_t = 1 - \eta_t^*\lambda_n/\rho_t = 1- \lambda_n/\rho_t^* = r_t^*$. We see that the definition of the set $\mathcal{S}^*$ immediately follows.
\end{proof}
\end{theorem}
In the above Theorem, we have shown that the best convergence rate is achievable, by tuning the step size, as long as $\rho_t \in [\lambda_n, \lambda_{q_t +1}]$. In the following Corollary, we provide an optimal choice for the tuple $(\rho_t, \eta_t)$ with respect to the estimation error.
\begin{corollary}\label{th:corollaryEE}
Among the tuples $(\rho_t, \eta_t) \in \mathcal{S}^*$, the choice of the tuple $(\rho_t^*, 1)$, with $\rho_t^*$ defined in (\ref{eq:defRhoStar}), is optimal with respect to the estimation error $\|\boldsymbol{\theta}^{t+1}_{\rho_t, \eta_t} - \boldsymbol{\theta}^*\|$, for any $\x^t$ and for any $t$, in the sense that
\begin{equation*}
    \|\boldsymbol{\theta}^{t+1}_{\rho_t^*,1}-\boldsymbol{\theta}^*\| \leq \|\boldsymbol{\theta}^{t+1}_{\rho_t,\eta_t}-\boldsymbol{\theta}^*\|, \ \ \forall{(\eta_t, \rho_t)} \in \mathcal{S}^*
\end{equation*}
\begin{proof}
Define $\mathbf{B}_t^* := (\mathbf{I} - (\hat{\boldsymbol{\Lambda}}_{t}(\rho_t^*, q_t))^{-1}\boldsymbol{\Lambda}) = diag(0, ...,0, 1-\lambda_{q_t+1}/\rho_t^*, ..., 1 - \lambda_n/\rho_t^*)$. For $\rho_t \neq \rho_t^*$, with $\rho_t \in [\lambda_{n}, \lambda_{q_t+1}]$, define $\mathbf{B}_t := (\mathbf{I} - \eta_t^*(\hat{\boldsymbol{\Lambda}}_{t}(\rho_t, q_t))^{-1}\boldsymbol{\Lambda}) = diag(1-\eta_t^*, ..., 1-\eta_t^*, 1-\lambda_{q_t+1}/\rho_t^*, ..., 1 - \lambda_n/\rho_t^*) = \mathbf{B}^*_t + \delta\mathbf{B}_t$, with $\delta\mathbf{B}_t = diag(1-\eta_t^*, ..., 1-\eta_t^*, 0, ..., 0)$, where $\eta_t^*$ is defined in (\ref{eq:bestEta}) and $\rho_t^*$ in (\ref{eq:defRhoStar}). Now define $\mathbf{z}^t := \mathbf{V}^T(\boldsymbol{\theta}^t - \boldsymbol{\theta}^*)$ and $\mathbf{z}^{t+1}_{\rho_t, \eta_t} := \mathbf{V}^T(\boldsymbol{\theta}^{t+1}_{\rho_t,\eta_t}-\boldsymbol{\theta}^*)$, where $(\boldsymbol{\theta}^{t+1}_{\rho_t,\eta_t}-\boldsymbol{\theta}^*)$ is defined in (\ref{eq:defAlgUpdate}). We have
\begin{equation*}
\begin{aligned}
     \|\boldsymbol{\theta}^{t+1}_{\rho_t^*,1}-\boldsymbol{\theta}^*\|^2 = \|\mathbf{z}^{t+1}_{\rho_t^*,1}\|^2 = \|\mathbf{B}^*_t\mathbf{z}^t\|^2,
\end{aligned}
\end{equation*}
\begin{equation*}
    \|\boldsymbol{\theta}^{t+1}_{\rho_t,\eta_t^*}-\boldsymbol{\theta}^*\|^2 = \|\mathbf{B}_t\mathbf{z}^t\|^2 = \|\mathbf{B}^*_t\mathbf{z}^t\|^2 + \|\delta\mathbf{B}_t\mathbf{z}^t\|^2,
\end{equation*}
because the cross term is $2(\mathbf{B}^*_t\mathbf{z}^t)^T(\delta\mathbf{B}_t\mathbf{z}^t) = 0$. We see that, for any $t$ and for any $\boldsymbol{\theta}^t$, 
\begin{equation*}
     \|\boldsymbol{\theta}^{t+1}_{\rho_t^*,1}-\boldsymbol{\theta}^*\| \leq \|\boldsymbol{\theta}^{t+1}_{\rho_t,\eta_t^*}-\boldsymbol{\theta}^*\|
\end{equation*}
\end{proof}
\end{corollary}
We remark that the bound in (\ref{tightBound}) is tight for $r_t = r_t^*$. If $q_t$ increases, the convergence factor $r_t^*$ decreases until it becomes zero, when $q_t = n-1$, thus we can have convergence in a finite number of steps.
\begin{figure}[t!]
\centering
		\includegraphics[width=\columnwidth]{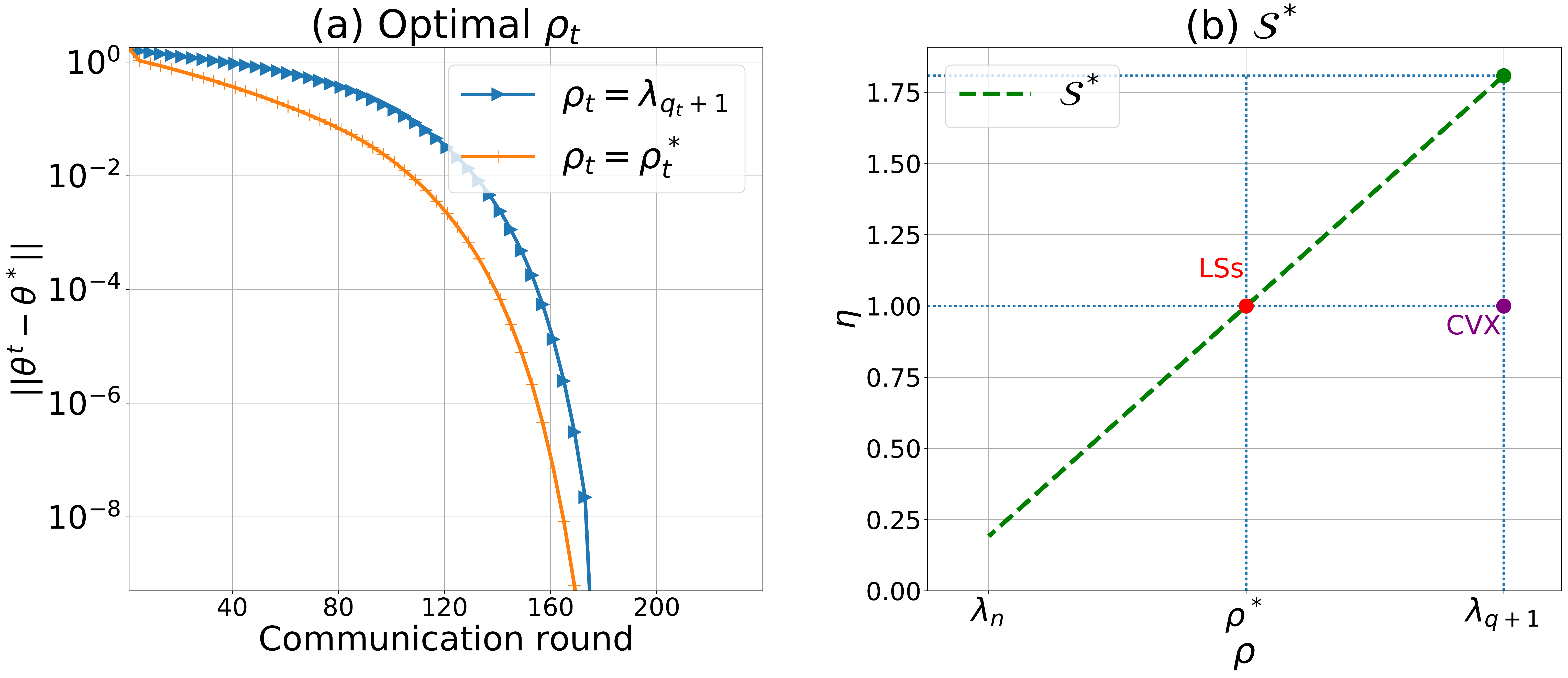}
		\caption{Performance comparison for different values of $\rho_t$. The best choice in terms of estimation error from Corollary \ref{th:corollaryEE} is compared against the choice that was proposed in \cite{MontanariWay}, that is $\rho_t = \lambda_{q_t+1}$. In (b), we show an example of the set $\mathcal{S}^*$, and outline two points: LSs in red is the choice of the tuple in $\mathcal{S}^*$ providing the optimal convergence factor in LSs, while the purple CVX which does not belong to $\mathcal{S}^*$, and it is the choice which we do in the scenario of FL with convex cost}
		\label{fig:rhoComp}
\end{figure}
\subsection{Federated least squares}
We now consider the FL scenario described in section \ref{sec:probForm}, in which $M$ agents keep their local data and share optimization parameters to contribute to the learning algorithm.
\begin{assumption}
In the rest of the paper, we assume that each agent has the same amount of data samples, $N_i = \frac{N}{M}$. This allows us to express global functions, such as the gradient, as the arithmetic mean of local functions (e.g., $\mathbf{g}_t = (1/M)\sum_{i = 1}^{M}\mathbf{g}_t^{(i)}$). 
\end{assumption}
This assumption is made only for notation convenience. It is straightforward to show that all the results are valid also for $N_i$ different for each $i$. To show it, it is sufficient to replace the arithmetic mean of local functions with the weighted average, weighting each local function with $p_i = N_i/N$. As an example, the global gradient would be written in the following way:
\begin{equation*}
    \mathbf{g}_t = \sum_{i = 1}^{M}p_i\mathbf{g}_t^{(i)}, p_i = N_i/N
\end{equation*}

In this subsection we introduce the algorithm for the LSs case (Algorithm~\ref{alg:1}), that is a special case of Algorithm~\ref{alg:2}, described in Sec. \ref{sec:generalConvex}, which is designed for a general convex cost. We refer to Algorithm 1 as SHED-LS and it works as follows: at iteration $t$, each device shares with the Master some of its local Hessian eigenvectors scaled by quantities related to the corresponding eigenvalues. The eigenvectors are shared \emph{incrementally}, where the order in which they are shared is given by the corresponding eigenvalues. For example, at $t = 1$ agent $i$ will start by sharing its first Hessian eigenvectors $\mathbf{v}_1^{(i)}, ..., \mathbf{v}_{q_1^{(i)}}^{(i)}$ according to its communication resources (CRs) and then will incrementally send up to $\mathbf{v}_{n-1}^{(i)}$ in the following iterations. To enable the approximation of the local Hessian via a limited number of eigenvectors using (\ref{eq:FullApprox}), each eigenvector is sent to the Master together with $\lambda_j^{(i)}$ and the parameter $\rho_t^{(i)}$.
\begin{algorithm}[t!]
\DontPrintSemicolon
  \KwInput{$\{\mathcal{D}_i\}_{i = 1}^{M} = \{(\mathbf{X}_i, \mathbf{Y}_i)\}_{i = 1}^{M}$, $t_{max}$, $\boldsymbol{\theta}^0$, $\mathcal{A} = \{agents\}$, $\mathcal{I} = \{1\}$}
  \KwOutput{$\x^{t_{max}}$}
  \For {$t \gets 1$ to $t_{max}$}
    {
        \For{$agent \ i \in \mathcal{A}$}
         {
            \when{Received $\x^t$ from the Master}{
            \If{$t \in \mathcal{I}$} 
            {
                compute $\mathbf{H}_{LS}^{(i)} = \nabla^2 f^{(i)}(\boldsymbol{\theta}^t) = \mathbf{X}_i\mathbf{X}_i^T$ \\
                $\{({\lambda}_{j}^{(i)}, {\mathbf{v}}_{j}^{(i)})\}_{j = 1}^{n} \gets eigendec(\mathbf{H}_{LS}^{(i)})$\tcp*{eigendecomposition}
                $q_{0}^{(i)} \gets 0$
            }
            compute $\mathbf{g}_t^{(i)} = \nabla f^{(i)}(\boldsymbol{\theta}^t)$\\
            set $d_t^{(i)}$ \tcp*{according to CRs}
            $q_t^{(i)} \gets q_{t-1}^{(i)} + d_t^{(i)}$ \tcp*{increment}
            ${\rho}_t^{(i)} \gets ({\lambda}_{q_t^{(i)}+1}^{(i)} + {\lambda}_{n}^{(i)})/2$\tcp*{approximation parameter}
            $U_t^{(i)} \gets \{\{{\mathbf{v}}_{j}^{(i)}, \lambda_j^{(i)}\}_{j = q^{(i)}_{t-1}+1}^{q^{(i)}_t}, \mathbf{g}_t^{(i)}, {\rho}_t^{(i)}\}$\\
            Send $U_t^{(i)}$ to the Master.
         }}
         \hrulefill\\
         At the Master:\\
         \when{Received $U_t^{(i)}$ from all agents}{
         compute $\hat{\mathbf{H}}_t^{(i)},\ \forall{i}$.\tcp*{see (\ref{eq:locLSHess})}
         Compute $\hat{\mathbf{H}}_t$ (as in eq. (\ref{eq:HTilde})) and
         $\mathbf{g}_t$.\\
         Perform Newton-type update (\ref{eq:qN-update}) with $\eta_t = 1$.\\
         Broadcast $\boldsymbol{\theta}^{t+1}$ to all agents.
         }
    }
\caption{Federated Least Squares - SHED-LS}\label{alg:1}
\end{algorithm}
The Master averages the received information to obtain an estimate of the global Hessian as follows:
\begin{equation}\label{eq:HTilde}
    {\hat{\mathbf{H}}}_t = \frac{1}{M}\sum_{i = 1}^{M}{\hat{\mathbf{H}}}_{t}^{(i)},
\end{equation}
where $\hat{\mathbf{H}}_t^{(i)}$ is
\begin{equation}\label{eq:locLSHess}
    \hat{\mathbf{H}}_{t}^{(i)} := \hat{\mathbf{H}}^{(i)}(\rho_t^{(i)}, q_t^{(i)}) =  \sum_{j = 1}^{q_t^{(i)}}(\lambda_j^{(i)} - \rho_t^{(i)}){\mathbf{v}}_j^{(i)}{{\mathbf{v}}_j^{(i)T}} + \rho_t^{(i)} \mathbf{I},
\end{equation}
in which $q_t^{(i)}$ is the number of the local Hessian eigenvectors-eigenvalues pairs that agent $i$ has already sent to the Master at iteration $t$. We denote by $d_t^{(i)}$ the \emph{increment}, meaning the number of eigenvectors that agent $i$ can send to the Master at iteration $t$. Given the results shown in Section \ref{sec:centralLSs}, in this Section we fix the local approximation parameter to be $\rho_t^{(i)} = \rho_t^{(i)*} = ({\lambda}_{q_t^{(i)}+1}^{(i)} + {\lambda}_{n}^{(i)})/2$ and the step size to be $\eta_t = 1$. The following results related to the convergence rate allow the value $q_t^{(i)}$ to be different for each agent $i$, so we define
\begin{equation*}
    \mathbf{q}_t = [q_t^{(1)}, ..., q_t^{(M)}]^{T}, \ {q}_t^{(i)} \in \{0, ...,n\} \ \ \forall{i\in\{1, ..., M\}}
\end{equation*}
By construction, the matrix $\hat{\mathbf{H}}_t$ is positive definite, being the sum of positive definite matrices, implying that $-\mathbf{p}_t = -\hat{\mathbf{H}}_t^{-1}\mathbf{g}_t$ is a descent direction.
\begin{theorem}\label{Th:theorem_1}
Consider the problem in (\ref{eq:sumOfConvexFunctions}) in the least squares case. Given $\hat{\mathbf{H}}_t$ defined in (\ref{eq:HTilde}), the update rule defined in (\ref{eq:qN-update}) is such that, for $\rho_t^{(i)} = (\lambda_{q_t^{(i)}+1}^{(i)} + \lambda_{n}^{(i)})/2$ and $\eta_t = 1$:
\begin{equation}\label{eq:linDecBound}
    \|\boldsymbol{\theta}^{t+1} - \boldsymbol{\theta}^*\| \leq c_{t}\|\boldsymbol{\theta}^t - \boldsymbol{\theta}^*\|,
\end{equation}
with $c_{t} = (1 - \bar\lambda_n/\Bar{\rho}_t)$ and
\begin{equation}\label{eq:defsOfRhoLBar}
    \Bar{\rho}_t := \bar{\rho}(\mathbf{q}_t) = \frac{1}{M}\sum_{i = 1}^{M}\rho_t^{(i)}, \ \ \Bar{\lambda}_n = \frac{1}{M}\sum_{i = 1}^{M}\lambda_n^{(i)}.
\end{equation}
If Algorithm 1 is applied, $c_{t+1} \leq c_{t} \ \forall{t}$, and, if for all $i$ it holds that $q_{t'}^{(i)} = {n-1}$, at some iteration $t'$ $c_{t'} = 0$.
\begin{proof}
Fix $\eta_t = 1$ in (\ref{eq:DT_system}),
\begin{equation*}
    \boldsymbol{\theta}^{t+1} - \boldsymbol{\theta}^* = \mathbf{A}_t({\x}^{t} - {\x}^*) = (\mathbf{I} - \hat{\mathbf{H}}_t^{-1}\mathbf{H}_{LS})({\x}^t - {\x}^*).
\end{equation*}
We have that
\begin{equation*}
    \begin{aligned}
         \|{\x}^{t+1} - {\x}^*\| &\leq \|\mathbf{I} - \hat{\mathbf{H}}_t^{-1}\mathbf{H}_{LS}\|\|{\x}^{t} - {\x}^*\| \\&
    \leq \|\hat{\mathbf{H}}_t^{-1}\|\|\hat{\mathbf{H}}_t - \mathbf{H}_{LS}\|\|{\x}^{t} - {\x}^*\|\\&
    \leq\frac{(\Bar{\rho}_t - \Bar{\lambda}_n)}{\Bar{\rho}_t}\|{\x}^{t} - {\x}^*\|
    \end{aligned}
\end{equation*}
The last inequality follows from two inequalities: (i) $\|\hat{\mathbf{H}}_t^{-1}\| \leq 1/\Bar{\rho}_t
$ and (ii) $\|\hat{\mathbf{H}}_t - \mathbf{H}_{LS}\| \leq \Bar{\rho}_t - \Bar{\lambda}_n$. \newline
(i) holds because $\|\hat{\mathbf{H}}_t^{-1}\| = (\lambda_{min}(\hat{\mathbf{H}}_t))^{-1}$, and $\lambda_{min}(\hat{\mathbf{H}}_t) \geq \Bar{\rho}_t$, thus implying $\|\hat{\mathbf{H}}_t^{-1}\| \leq 1/\Bar{\rho}_t
$. \newline
(ii) follows recalling that $\mathbf{H}_{LS} = \frac{1}{M}\sum_{i = 1}^{M}\mathbf{H}^{(i)}_{LS}$, with $\mathbf{H}^{(i)}_{LS}$ the local Hessian at agent $i$. We have
\begin{equation*}
    \|\hat{\mathbf{H}}_t - \mathbf{H}_{LS}\| = \frac{1}{M}\|\sum_{i = 1}^{M}(\hat{\mathbf{H}}_t^{(i)} - \mathbf{H}^{(i)}_{LS})\| \leq \frac{1}{M}\sum_{i = 1}^{M}\|\hat{\mathbf{H}}_t^{(i)} - \mathbf{H}^{(i)}_{LS}\|.
\end{equation*}
Being $\hat{\mathbf{H}}_t^{(i)} - \mathbf{H}^{(i)}_{LS}$ symmetric, it holds that
\begin{equation*}
    \|\hat{\mathbf{H}}_t^{(i)} - \mathbf{H}^{(i)}_{LS}\| = \max_j|\lambda_j(\hat{\mathbf{H}}_t^{(i)} - \mathbf{H}^{(i)}_{LS})| = \rho_t^{(i)} - \lambda_n^{(i)},
\end{equation*}
where the last equality holds because
\begin{equation}\label{eq:individH_i}
    \hat{\mathbf{H}}_t^{(i)} - \mathbf{H}^{(i)}_{LS} = \mathbf{V}^{(i)}(\hat{\boldsymbol{\Lambda}}_t^{(i)} - \boldsymbol{\Lambda}_t^{(i)})\mathbf{V}^{(i)T}
\end{equation}
where 
\begin{equation}\label{eq:theLambdas}
\begin{aligned}
    &\hat{\boldsymbol{\Lambda}}_t^{(i)} = diag(\lambda_1^{(i)},...,\lambda_{q_t^{(i)}}^{(i)}, \rho_t^{(i)}, ..., \rho_t^{(i)}),\\&
    \boldsymbol{\Lambda}_t^{(i)} = diag(\lambda_1^{(i)}, ...,\lambda_{q_t^{(i)}}^{(i)}, \lambda_{q_t^{(i)}+1}^{(i)}, ... \lambda_n^{(i)}),
\end{aligned}
\end{equation} and because $\rho_t^{(i)} =  (\lambda_{q_t^{(i)}+1}^{(i)} + \lambda_{n}^{(i)})/2$. 
\end{proof}
\end{theorem}
This theorem shows that our approach in the LSs case provides convergence in a finite number of iterations, if $q_t^{(i)}$ keeps increasing through time for each agent $i$. Indeed, as in the centralized case, if $q_t^{(i)}$ increases for all $i$, the factor $c_t$ decreases until it becomes zero. Furthermore, each agent is free to send at each iterations an arbitrary number of eigenvector-eigenvalue pairs, according to its CRs, and by doing so it can improve the convergence rate.


%% file: JMLR/section_4.tex
We want to extend the analysis and algorithm presented in the previous sections of the paper to a general convex cost $f(\x^t)$. With respect to the proposed approach, the general convex case requires special attention for two main reasons: (i) the update rule defined in (\ref{eq:qN-update}) requires tuning of the step-size $\eta_t$, usually via backtracking line search, and (ii) the Hessian matrix is in general a function of the parameter $\x$. 
In this section, still focusing on the least squares case, we provide some results that are instrumental to the analysis of the general convex case. 
\subsection{Backtracking line search for step size tuning}\label{subsec:lineSearch}
We recall the well-known Armijo-Goldstein condition for accepting a step size $\eta_t$ via backtracking line search:
\begin{equation}\label{eq:Armijo}
	f(\boldsymbol{\theta}^{t} - \eta_t \mathbf{p}) \leq f(\boldsymbol{\theta}^{t}) - \alpha \eta_t \mathbf{p}^T\mathbf{g}_t,
\end{equation}
where $\alpha \in (0, 1/2)$. The corresponding line search algorithm is the following:
\begin{algorithm}
	\DontPrintSemicolon
	\KwInput{$\alpha \in (0, 1/2)$, $\beta \in (0, 1)$, $\x^t$, $\mathbf{p}_t$, $\mathbf{g}_t$, $f$}
	\KwOutput{$\bar{\eta}_t$}
	$\eta_t^0 \gets 1$, $k \gets 0$\\
	\While{$f(\boldsymbol{\theta}^{t} - \eta_t^{(k)} \mathbf{p}) > f(\boldsymbol{\theta}^{t}) - \alpha \eta_t^{(k)} \mathbf{p}^T\mathbf{g}_t$}
	{
		$k \gets k + 1$\\
		$\eta_t^{(k)} = \beta\eta_t^{(k-1)}$
	}
	$\bar{\eta}_t = \eta_t^{(k)}$
	\caption{Backtracking line search algorithm}\label{alg:lineSearch}
\end{algorithm}
\begin{lemma}\label{lemma:ArmijoLSs}
	Consider the problem in (\ref{eq:sumOfConvexFunctions}) in the least squares case. Let $\mathbf{p}_t = \hat{\mathbf{H}}_t^{-1}\mathbf{g}_t$, with $\hat{\mathbf{H}}_t$ defined in (\ref{eq:HTilde}). A sufficient condition for a step size $\eta_t$ to satisfy Armijo-Goldstein condition (\ref{eq:Armijo}), for any $\alpha \in (0, 1/2)$, is
	\begin{equation}\label{eq:conditionSS}
		\eta_t = \min_{i = 1, ..., M}{\frac{\rho_t^{(i)}}{\lambda_{q^{(i)}_t+1}^{(i)}}}
	\end{equation}
	\begin{proof} The quadratic cost in $\boldsymbol{\theta}^t$ can be written as
		\begin{equation*}\label{eq:cost}
			f(\boldsymbol{\theta}^t) = f(\boldsymbol{\theta}^*) +  \bar{f}(\boldsymbol{\theta}^t).
		\end{equation*}
		with $\bar{f}(\boldsymbol{\theta}^t) = \frac{1}{2}(\boldsymbol{\theta}^t - \boldsymbol{\theta}^*)^T\mathbf{H}_{LS}(\boldsymbol{\theta}^t - \boldsymbol{\theta}^*)$. Given that $f(\boldsymbol{\theta}^*)$ does not depend on $\boldsymbol{\theta}^t$, we can focus on $\bar{f}(\boldsymbol{\theta}^t)$.\newline
		We have that
		\begin{equation}\label{eq:GetSuffCond}
			\begin{aligned}[b]
				\bar{f}(\boldsymbol{\theta}^t - \eta_t\mathbf{p}_t) &= \frac{1}{2}(\boldsymbol{\theta}^t -  \eta_t\mathbf{p}_t - \boldsymbol{\theta}^*)^T\mathbf{H}_{LS}(\boldsymbol{\theta}^t - \eta_t\mathbf{p}_t - \boldsymbol{\theta}^*)\\&
				\stackrel{(1)}{=} \bar{f}(\boldsymbol{\theta}^t) + \frac{1}{2} \eta_t^2\mathbf{p}_t^T \mathbf{H}_{LS}\mathbf{p}_t - \eta_t\mathbf{p}_t^T  \mathbf{g}_t\\&
				\stackrel{(2)}{=} \bar{f}(\boldsymbol{\theta}^t) - \eta_t\mathbf{p}_t^T(\hat{\mathbf{H}}_t - \eta_t\frac{\mathbf{H}_{LS}}{2})\mathbf{p}_t
			\end{aligned}
		\end{equation}
		where we have used identity $\mathbf{H}_{LS}(\boldsymbol{\theta}_t-\boldsymbol{\theta}^*) = \mathbf{g}_t$ and the fact that $ \mathbf{p}_t^T  \mathbf{g}_t =  \mathbf{p}_t^T\hat{\mathbf{H}}_t\hat{\mathbf{H}}_t^{-1}\mathbf{g}_t = \mathbf{p}_t^T\hat{\mathbf{H}}_t\mathbf{p}_t$ to get equality (1) and (2), respectively. We see that if 
		\begin{equation}\label{eq:SuffCondArmijo}
			\hat{\mathbf{H}}_t - \eta_t\mathbf{H}_{LS}/2 \geq \hat{\mathbf{H}}_t/2,    
		\end{equation}
		then Armijo-Goldstein condition (\ref{eq:Armijo}) is satisfied. Indeed, in that case,
		\begin{equation*}
			\begin{aligned}
				\bar{f}(\boldsymbol{\theta}^t) - \eta_t\mathbf{p}_t^T(\hat{\mathbf{H}}_t - \eta_t\frac{\mathbf{H}_{LS}}{2})\mathbf{p}_t &\leq \bar{f}(\boldsymbol{\theta}^t) - \frac{1}{2}\eta_t\mathbf{p}_t^T\hat{\mathbf{H}}_t\mathbf{p}_t\\& \leq \bar{f}(\boldsymbol{\theta}^t) - \alpha\eta_t\mathbf{p}_t^T\mathbf{g}_t.
			\end{aligned}
		\end{equation*}
		So, we need to find a sufficient condition on $\eta_t$ for (\ref{eq:SuffCondArmijo}) to be true. We see that (\ref{eq:SuffCondArmijo}) is equivalent to $\hat{\mathbf{H}}_t - \eta_t\mathbf{H}_{LS} \geq 0$. We have $\hat{\mathbf{H}}_t - \eta_t\mathbf{H}_{LS} = \frac{1}{M}\sum_{i = 1}^{M}\mathbf{V}^{(i)}(\hat{\boldsymbol{\Lambda}}_t^{(i)} - \eta_t\boldsymbol{\Lambda}_t^{(i)})\mathbf{V}^{(i)T}$, and, $\forall{i}$, all the elements of the diagonal matrix $\hat{\boldsymbol{\Lambda}}_t^{(i)} - \eta_t\boldsymbol{\Lambda}_t^{(i)}$ are positive if $\eta_t \leq \rho^{(i)}_t/\lambda_{q_t^{(i)}+1}^{(i)}$ (see Eq. (\ref{eq:theLambdas})), and we see that the choice (\ref{eq:conditionSS}) provides a sufficient condition to satisfy (\ref{eq:SuffCondArmijo}), from which we can conclude. 
	\end{proof}
\end{lemma}
\begin{corollary}\label{corollary:ArmijoLSs}
	In the least squares case (Algorithm~\ref{alg:1}), choosing $\rho_t^{(i)} = \rho_t^{(i)*}, \ \forall{i}$, Armijo backtracking line search (Algorithm \ref{alg:lineSearch}) would choose a step size $\eta_t \geq \frac{1}{2}$. Choosing $\rho_t^{(i)} = \lambda_{q_t^{(i)} + 1}, \ \forall{i}$, Armijo backtracking line search would choose a step size $\eta_t = 1$.
	\begin{proof}
		The proof is straightforward from Eqn. (\ref{eq:conditionSS}) of Lemma \ref{lemma:ArmijoLSs}.
	\end{proof}
\end{corollary}
\begin{remark}
For the choice $\rho^{(i)} = \rho^{(i)*}$, the Armijo-backtracking might not choose a step size $\eta_t = 1$ even for arbitrarily small $\alpha < 0.5$. Indeed, we can easily build a counter-example in the centralized case considering $\Hat{\mathbf{H}}_t = \Hat{\boldsymbol{\Lambda}}_t = diag(\lambda_1, ..., \lambda_q, \rho^*, ..., \rho^*)$, a gradient $\mathbf{g}_t$ such that $\mathbf{p}_t = {\mathbf{\Hat{H}}}_t^{-1}\mathbf{g}_t = [0, ...,0, 1, 0, ..., 0]^\top$ (e.g., $\mathbf{g}_t = [0, ...0, \lambda_{q_t+1}, 0, ..., 0]^\top$). We see that with $\eta_t = 1$, Eq. (\ref{eq:GetSuffCond}) becomes $f({\x}^{t+1}) = f({\x}^t) - \lambda_n/2$ and, in order to be satisfied, the Armijo condition would require $\alpha \leq \frac{\lambda_n}{\lambda_{q+1} + \lambda_n}$, where the right hand side can become arbitrarily small depending on the eigenspectrum. 
\end{remark}
The results of Lemma \ref{lemma:ArmijoLSs} and of Corollary \ref{corollary:ArmijoLSs} and the counter-example of the above Remark are important for the design of the algorithm in the general convex case. Indeed, as illustrated in the next Section (Section~\ref{sec:generalConvex}), a requirement for the theoretical results on the convergence rate is that the step size becomes equal to one, which is not guaranteed by the Armijo backtracking line search, even when considering the least squares case, if $\rho_t^{(i)} < \lambda_{q_t^{(i)} + 1}^{(i)}$. For this reason, the algorithm in the general convex case is designed with $\rho_t^{(i)} = \lambda_{q_t^{(i)} + 1}^{(i)}$
\subsection{Algorithm with periodic renewals}\label{subsec:Periodic}
Now, we introduce a variant of Algorithm \ref{alg:1} that is instrumental to study the convergence rate of the proposed algorithm in the general convex case.
The variant is Algorithm \ref{alg:variantOf1}. The definition of $\mathcal{I} = \{1, T, 2T, ...\}$ implies that every $T$ iterations the incremental strategy is restarted from the first EEPs of $\mathbf{H}_{LS}$, in what we call a periodic \emph{renewal}.
\begin{algorithm}
	\DontPrintSemicolon
	In Algorithm \ref{alg:1}, substitute $\mathcal{I} = \{1\}$ with $\mathcal{I} = \{1, T, 2T, ...\}$, for some input parameter $T < n$.
	\caption{Variant of Algorithm \ref{alg:1}, SHED-LS-periodic}\label{alg:variantOf1}
\end{algorithm}
Differently from Algorithm \ref{alg:1}, Algorithm \ref{alg:variantOf1} can not guarantee convergence in a finite number of steps, because $T < n$ and thus it could be that $c_t > 0, \ \forall{t}$ (see Theorem \ref{Th:theorem_1}). We study the convergence rate of the algorithm by focusing on upper bounds on the Lyapunov exponent \citep{LyapunovExp} of the discrete-time dynamical system ruled by the descent algorithm. The Lyapunov exponent characterizes the rate of exponential (linear) convergence and it is defined as the positive constant  $a_* > 0$ such that, considering $h(\x^t) :=  (\x^t - \x^*)a^{-t}$, if $a > a_*$ then $h(\x^t)$ vanishes with $t$, while if $a < a_*$, for some initial condition, $h(\x^t)$ diverges. The usual definition of Lyapunov exponent for discrete-time linear systems \citep[see][]{LyapunovDefinition} is, considering the system defined in (\ref{eq:DT_system}),
\begin{equation}\label{def:LyapExp}
	a_* := \limsup_{t \rightarrow \infty}{\|\boldsymbol{\Psi}_t\|^{1/t}}, \ \boldsymbol{\Psi}_t = \mathbf{A}_1\cdots \mathbf{A}_t.
\end{equation}
From (\ref{eq:linDecBound}), we have that, for each $k$, $\|\mathbf{A}_k\| \leq c_k = (1 - \bar{\lambda}_n/\bar{\rho}_k)$. This implies that, defining 
\begin{equation}\label{eq:LyapExp}
	\begin{aligned}
		a_t :&= (\prod_{k = 1}^{t}{c_k})^{1/t},\\
		\bar{a} :&= \limsup_{t \rightarrow +\infty}{a_t},
	\end{aligned}
\end{equation}
it is $a_* \leq \bar{a}$. The following Lemma formalizes this bound and provides an upper bound on the Lyapunov exponent obtained by applying Algorithm \ref{alg:variantOf1}.
\begin{lemma}\label{lemma:LyapLSs}
	Let $c_k = (1 - \bar{\lambda}_n/\bar{\rho}_k)$ (see Eq. (\ref{eq:defsOfRhoLBar})). Applying Algorithm \ref{alg:variantOf1}, the Lyapunov exponent of system (\ref{eq:DT_system}) is such that
	\begin{equation*}
		a_* \leq \bar{a} = \limsup_{t \rightarrow +\infty}(\prod_{k = 1}^{t}{c_k})^{1/t}.
	\end{equation*}
	If $q_t^{(i)} = q_{t-1}^{(i)} + 1, \ \forall{i, t},$
	\begin{equation}\label{eq:LyapBoundLSs}
		a_* \leq \bar{a}_T := (\prod_{k = 1}^{T}c_k)^{1/T}.
	\end{equation}
	where $\bar{a}_T$ is such that $\bar{a}_{T+1} \leq \bar{a}_{T}$ and $\bar{a}_n = 0$.
	\begin{proof}
		We see that $a_* \leq \bar{a}$ from definition (\ref{def:LyapExp}), because for each $k$, $\|\mathbf{A}_k\| \leq c_k = (1 - \bar{\lambda}_n/\bar{\rho}_k)$. Now, we show that, if $q_t^{(i)} = q_{t-1}^{(i)} + 1, \ \forall{i, t}$, then $\bar{a} = \bar{a}_T$. To show this latter inequality, let us consider the logarithm of the considered values, so we prove $\log{\bar{a}} = \log{\bar{a}_T}$. Indeed, if $q_t^{(i)} = q_{t-1}^{(i)} + 1$ and applying Algorithm \ref{alg:variantOf1}, $c_k$ is a periodic sequence of the index $k$ ($c_{k+T} = c_k$):
		\begin{equation}\label{eq:periodicObtain}
			\begin{aligned}
				\log{\bar{a}} &= \limsup_t{\frac{1}{t}\sum_{k = 1}^{t}\log{c_k}}\\& = \limsup_{R}{\frac{1}{RT + T'}(R\sum_{k = 1}^T\log{c_k} + \sum_{k = 1}^{T'}\log{c_k}})\\& = \frac{1}{T}\sum_{k = 1}^{T}\log{c_k} = \log{\bar{a}_T}
			\end{aligned}
		\end{equation}
		where $R = \floor{t/T}$ and $T' = t - RT$.
	\end{proof}
\end{lemma}

%% file: JMLR/section_5.tex
Given the previous analysis and theoretical results for linear regression with quadratic cost, we are now ready to illustrate our Newton-type algorithm (Algorithm \ref{alg:2}) for general convex FL problems, of which Algorithm 1 is a special case. We refer to the this general version of the algorithm simply as SHED.
\begin{algorithm}[t]
	\DontPrintSemicolon
	\KwInput{$\{\mathcal{D}_i\}_{i = 1}^{M}$, $t_{max}$, $\mathcal{I}$, $\boldsymbol{\theta}^0$, $\mathcal{A} = \{agents\}$}
	\KwOutput{$\x^{t_{max}}$}
	\For {$t \gets 0$ to $t_{max}$}
	{
		\For{$agent \ i \in \mathcal{A}$}
		{
			\when{Received $\x^t$ from the Master}{
				\If{$t \in \mathcal{I}$} 
				{
					$k_t \gets t$ \\
					compute $\mathbf{H}_t^{(i)} = \nabla^2 f^{(i)}(\boldsymbol{\theta}^t)$ \tcp*{renewal}
					$\{(\hat{\lambda}_{j, t}^{(i)}, \hat{\mathbf{v}}_{j, t}^{(i)})\}_{j = 1}^{n} \gets eigendec(\mathbf{H}_t^{(i)})$\tcp*{eigendecomposition}
					$q_{t-1}^{(i)} \gets 0$
				}
				compute $\mathbf{g}_t^{(i)} = \mathbf{g}^{(i)}(\boldsymbol{\theta}^t)= \nabla f^{(i)}(\boldsymbol{\theta}^t)$\\
				set $d_t^{(i)}$ \tcp*{according to CRs}
				$q_t^{(i)} \gets q_{t-1}^{(i)} + d_t^{(i)}$ \tcp*{increment}
				$\hat{\rho}_t^{(i)} \gets \hat{\lambda}_{q_t^{(i)}+1, t}^{(i)}$ \tcp*{approximation parameter}
				$U_t^{(i)} \gets \{\{{\hat{\mathbf{v}}}_{j, t}^{(i)}, \hat{\lambda}_{j, t}^{(i)}\}_{j = q^{(i)}_{t-1}+1}^{q^{(i)}_t}, \mathbf{g}_t^{(i)}, {\hat{\rho}}_t^{(i)}\}$ \tcp*{see (\ref{eq:vtilde}) and (\ref{eq:locCxHess})}
				Send $U_t^{(i)}$ to the Master.
		}}
		\hrulefill\\
		At the Master:\\
		\when{Received $U_t^{(i)}$ from all agents}{
			compute $\hat{\mathbf{H}}_t^{(i)},\ \forall{i}$.\tcp*{see (\ref{eq:locCxHess})}
			Compute $\hat{\mathbf{H}}_t$ (as in eq. (\ref{eq:aggregation})) and
			$\mathbf{g}_t$.\\
			Get $\eta_t$ via federated Armijo backtracking line search.\\
			Perform Newton-type update (\ref{eq:qN-update}).\\
			Broadcast $\boldsymbol{\theta}^{t+1}$ to all agents.}
	}
	\caption{FL with convex cost - SHED}\label{alg:2}
\end{algorithm}
Since in a general convex problem the Hessian depends on the current parameter, $\boldsymbol{\theta}^t$, we denote by $\mathbf{H}(\boldsymbol{\theta}^t)$ the global Hessian at the current iterate, while we denote by $\hat{\mathbf{H}}_t$ the global approximation, defined similarly to (\ref{eq:HTilde}), with the difference that now eigenvalues and eigenvectors depend on the parameter for which the Hessian was computed. \\
The expression of $\hat{\mathbf{H}}_t$ thus becomes:
\begin{equation}\label{eq:aggregation}
	\hat{\mathbf{H}}_t = \frac{1}{M}\sum_{i = 1}^{M}\hat{\mathbf{H}}_t^{(i)}{(\boldsymbol{\theta}^{k_{t}^{(i)}})},
\end{equation}
where $k_t^{(i)} \leq t$ denotes the iteration in which the local Hessian of agent $i$ was computed. The parameter $\boldsymbol{\theta}^{k_t^{(i)}}$ is the parameter for which agent $i$ computed the local Hessian, that in turn is being used for the update  at iteration $t$.\\ 
The idea of the algorithm is to use previous versions of the Hessian rather than always recomputing it. This is motivated by the fact that as we approach the solution of the optimization problem, the second order approximation becomes more accurate and the Hessian changes more slowly. Hence, recomputing the Hessian and restarting the incremental approach provides less and less advantages as we proceed. From time to time, however, we need to re-compute the Hessian corresponding to the current parameter $\boldsymbol{\theta}^{t}$, because $\mathbf{H}(\boldsymbol{\theta}^{k_t})$ could have become too different from $\mathbf{H}(\boldsymbol{\theta}^t)$. As in Section~\ref{subsec:Periodic}, we call this operation a \emph{renewal}. We denote by $\mathcal{I}$ the set of iteration indices at which a renewal takes place. In principle, each agent could have its own set of renewal indices, and decide to recompute the Hessian matrix independently. In this work, we consider for simplicity that the set $\mathcal{I}$ is the same for all agents, meaning that all agents use the same parameter for the local Hessian computation, i.e., $k_t^{(i)} = k_t, \ \forall{i}$. At the end of this section we describe heuristic strategies to choose $\mathcal{I}$ with respect to the theoretical analysis. We remark that in the case of a quadratic cost, in which the Hessian is constant, one chooses $\mathcal{I} = \{1\}$, and so Algorithm \ref{alg:1} is a special case of Algorithm \ref{alg:2}.
\newline The eigendecomposition can be applied to the local Hessian as before, we define
\begin{equation}\label{eq:defsConvex}
	\hat{\mathbf{v}}_{j, t}^{(i)} = \mathbf{v}_j^{(i)}(\boldsymbol{\theta}^{k_{t}}), \ \  \hat{\lambda}_{j, t}^{(i)} = \lambda_j^{(i)}(\boldsymbol{\theta}^{k_{t}}).
\end{equation}
For notation convenience we also define
\begin{equation}\label{eq:vtilde}
	\Tilde{\mathbf{v}}_{j, t}^{(i)} = (\hat{\lambda}_{j, t}^{(i)} - \hat{\rho}_t^{(i)})^{1/2}\hat{\mathbf{v}}_{j, t}^{(i)},
\end{equation}
The theoretical results on the convergence rate in this Section require that $\hat{\mathbf{H}}^{(i)}({\x}) \geq \mathbf{H}^{(i)}(\x), \forall{\x}$, which in turn requires, defining $\hat{\rho}_t^{(i)} = {\rho}_t^{(i)}(\x^{k_t})$, that $\hat{\rho}_t^{(i)} \geq \hat{\lambda}^{(i)}_{q_t^{(i)} +1, t}$. Given the results of Theorem \ref{lemma_1} on the range of the approximation parameter $\rho_t$ achieving the best convergence rate in the centralized least squares case, we set:
\begin{equation}\label{eq:rhoConvex}
	\hat{\rho}_t^{(i)} = \hat{\lambda}_{q_t+1, t}^{(i)}.
\end{equation}
The local Hessian can be approximated as
\begin{equation}\label{eq:locCxHess}
	\hat{\mathbf{H}}_t^{(i)}(\boldsymbol{\theta}^{k_{t}}) = \sum_{j = 1}^{q_t^{(i)}}{\Tilde{\mathbf{v}}_{j, t}^{(i)}\Tilde{\mathbf{v}}_{j, t}^{(i)T}} + \hat{\rho}_t^{(i)}\mathbf{I}.
\end{equation}
Clearly, it still holds that $\hat{\mathbf{H}}_t \geq \Bar{\rho}_t\mathbf{I}, \ \forall{t}$, where $\bar{\rho}_t = \frac{1}{M}\sum_{i = 1}^{M}\hat{\rho}_t^{(i)}$. Furthermore, it is easy to see that $\hat{\mathbf{H}}_t \leq K\mathbf{I}$, with $K$ the smoothness constant of $f$. We assume to use Armijo backtracking condition, that is recalled in Algorithm \ref{alg:lineSearch}.

\begin{theorem}\label{corollary:convergence}
	For any initial condition, Algorithm 2 ensures convergence to the optimum, i.e.,
	\begin{equation*}
		\lim_{t\rightarrow +\infty}\|\boldsymbol{\theta}_t - \boldsymbol{\theta}^*\| = 0.
	\end{equation*}
	\begin{proof}
	See Appendix A.
\end{proof}\vspace{-0.4cm}
\end{theorem}

Now, we provide results related to the convergence rate of the algorithm. In order to prove the following results, we need to impose some constraints on the renewal indices set $\mathcal{I}$. Specifically,
\begin{assumption}\label{th:AssDelayBound}
	Denoting $\mathcal{I} = \{C_j\}_{j\in \mathbb{N}}$, there exists a finite positive integer $\bar{l}$ such that $C_j \leq C_{j-1} + \bar{l}, \ \forall{j}$. This is equivalent to state that, writing $k_t = t - \tau$, the ‘delay’ $\tau$ is bounded.
\end{assumption}

The next Theorem provides a bound describing the relation between the convergence rate and the increments of outdated Hessians.
\begin{theorem}\label{Th:theorem_2}
	Applying SHED (Algorithm \ref{alg:2}), for any iteration $t$, it holds that:
	\begin{equation}\label{eq:convBound}
		\|\boldsymbol{\theta}^{t+1} - \boldsymbol{\theta}^*\| \leq c_{1,t}\|\boldsymbol{\theta}^t - \boldsymbol{\theta}^*\| + c_{2,t}\|\boldsymbol{\theta}^t - \boldsymbol{\theta}^*\|^2
	\end{equation}
	where, defining $\bar\lambda_{n, t} = \frac{1}{M}\sum_{i = 1}^{M}\hat{\lambda}_{{n}, t}^{(i)}$ and $\Bar{\rho}_t = \frac{1}{M}\sum_{i = 1}^{M}\hat{\rho}_t^{(i)}$ (see (\ref{eq:defsConvex}) and (\ref{eq:rhoConvex})),
	\begin{equation}
		\begin{aligned}
			c_{1,t} &= (1 - \frac{\bar\lambda_{n, t}}{\Bar{\rho}_t}) + \frac{L}{\Bar{\rho}_t}\|\boldsymbol{\theta}^t - \boldsymbol{\theta}^{k_t}\| +  (1-\eta_t)\frac{\|{\mathbf{H}}(\boldsymbol{\theta}^t)\|}{\bar{\rho}_t}, \ \ \\ c_{2,t} &= \frac{\eta_tL}{2\Bar{\rho}_t}.
		\end{aligned}
	\end{equation}
	\begin{proof}
		The beginning of the proof follows from the proof of Lemma 3.1 in \cite{MontanariWay}, (see page 18). In particular, we can get the same inequality as (A.1) in \cite{MontanariWay} (the $\mathbf{Q}^t$ here is $\hat{\mathbf{H}}_t$) with the difference that since we are not sub-sampling, we have (using the notation of \cite{MontanariWay}) $S = [n]$. The following inequality holds:
		\begin{equation*}
			\begin{aligned}
				\|\boldsymbol{\theta}^{t+1} - \boldsymbol{\theta}^*\| &\leq \|\boldsymbol{\theta}^t - \boldsymbol{\theta}^*\|\|I - \eta_t\hat{\mathbf{H}}_t^{-1}\mathbf{H}(\boldsymbol{\theta}^t)\|\\& + \eta_tL\frac{\|\hat{\mathbf{H}}_t^{-1}\|}{2} \|\boldsymbol{\theta}^{t} - \boldsymbol{\theta}^*\|^2.
			\end{aligned}
		\end{equation*}
			Note that, as we have shown in the proof of Theorem \ref{Th:theorem_1}, it holds $\|\hat{\mathbf{H}}_t^{-1}\| \leq 1/\bar{\rho}_t$. We now focus on the first part of the right hand side of the inequality:
		\begin{equation*}\label{eq:stepSizeBound}
			\begin{aligned}
				\|I - \eta_t\hat{\mathbf{H}}_t^{-1}\mathbf{H}(\boldsymbol{\theta}^t)\| &\leq \|\hat{\mathbf{H}}_t^{-1}\|(\|\hat{\mathbf{H}}_t - \mathbf{H}(\boldsymbol{\theta}^t)\| \\&+(1 - \eta_t){\|{\mathbf{H}}(\boldsymbol{\theta}^t)\|}) \\& \leq\frac{1}{\bar{\rho}_t}\|\hat{\mathbf{H}}_t - \mathbf{H}(\boldsymbol{\theta}^t)\| + \frac{(1 - \eta_t)}{\bar{\rho}_t}{\|{\mathbf{H}}(\boldsymbol{\theta}^t)\|}
			\end{aligned}
		\end{equation*}
		and focusing now on the first term of the right hand side of the last inequality
		\begin{equation*}
			\begin{aligned}
				&\frac{1}{\Bar{\rho}_t}(\|\hat{\mathbf{H}}_t - \mathbf{H}(\boldsymbol{\theta}^{k_t})\| + \|\mathbf{H}(\boldsymbol{\theta}^t) - \mathbf{H}(\boldsymbol{\theta}^{k_t})\|)\\\leq&
				\frac{1}{\Bar{\rho}_t}(\frac{1}{M}\sum_{i = 1}^M  \|\hat{\mathbf{H}}_t^{(i)}(\boldsymbol{\theta}^{k_{t}}) - \mathbf{H}^{(i)}(\boldsymbol{\theta}^{k_t})\| + L\|\boldsymbol{\theta}^{k_{t}} - \boldsymbol{\theta}^t\|)\\=&
				1 - \frac{\Bar{\lambda}_{n,t}}{\bar{\rho}_t} + \frac{L}{\bar{\rho}_t}\|\boldsymbol{\theta}^{k_{t}} - \boldsymbol{\theta}^t\|,
			\end{aligned}
		\end{equation*}
		where the last equality holds being $\|\hat{\mathbf{H}}_t^{(i)}(\boldsymbol{\theta}^{k_{t}}) - \mathbf{H}^{(i)}(\boldsymbol{\theta}^{k_t})\| =\hat{\rho}_t^{(i)} - \lambda_{n,t}^{(i)}$, which in turn is true given that $\hat{\rho}_t^{(i)} = \hat{\lambda}_{q_t^{(i)}+1, t}^{(i)}$.
	\end{proof}\vspace{-0.3cm}
\end{theorem}
The above theorem is a generalization of Lemma 3.1 in \cite{MontanariWay} (without sub-sampling). In particular, the difference is that (i) the dataset is distributed and (ii) an outdated Hessian matrix is used. 

\begin{theorem}\label{Th:newLinSupLin}
	Recall the definition of the average strong convexity constant $\bar{\kappa} = (1/M)\sum_{i = 1}^{M}\kappa_i$, with $\kappa_i$ the strong convexity constant of agent $i$. Let $K$ be the smoothness constant of $f$. The following results hold:
	
	\begin{enumerate}
		\item If
		\begin{equation}\label{eq:condStepSize}
			3\bar{\kappa}(M(t) + \|{\x}^t - {\x}^*\|) + K\|{\x}^t - {\x}^*\| \leq \frac{3\bar{\kappa}^2}{L}(1-2\alpha),
		\end{equation}
		and 
		\begin{equation}\label{eq:condContract}
			\frac{3}{2}L\|{\x}^t - {\x}^*\| + LM(t) \leq \bar{\kappa}
		\end{equation}
	with $M(t) = \max\{\|{\x}^t - {\x}^*\|, \|{\x}^{k_t} - {\x}^*\|\}$
		then SHED (Algorithm \ref{alg:2}) enjoys at least linear convergence.
		\item Define $\mathcal{X}_t := \{k \leq t: \bar{\rho}_k = \bar{\lambda}_{n, k}\}$. Let $|\mathcal{X}_t|$ denote the cardinality of $\mathcal{X}_t$. If $|\mathcal{X}_t| = 0$ then SHED enjoys linear convergence and the Lyapunov exponent of the estimation error can be upper bounded as
		\begin{equation}
			{a_*} \leq \limsup_t(\prod_{k = 1}^{t}{1 - \frac{\bar{\lambda}_n^{o}}{\bar{\rho}_{k}^{o}}})^{1/t}
		\end{equation}
	where $\bar{\lambda}_n^{o}$ and $\bar{\rho}_k^{o}$ are the average of the $n$-th eigenvalues and approximation parameters, respectively, computed at the optimum:
	\begin{equation*}
		\bar{\lambda}_n^{o} = \frac{1}{M}\sum_{i = 1}^{M}\lambda_n^{(i)}(\boldsymbol{\theta}^{*}), \ \ \bar{\rho}_k^{o} = \frac{1}{M}\sum_{i = 1}^{M}\rho_{k}^{(i)o},
	\end{equation*}
	with $\rho_{k}^{(i)o} = \lambda_{q_k^{(i)} + 1}^{(i)}(\boldsymbol{\theta}^{*})$.
	\item Let $|\mathcal{X}_t|$ denote the cardinality of $\mathcal{X}_t$. If $|\mathcal{X}_t| \geq t^{1/2}h(t)$, with $h(t)$ any function such that $h(t) \rightarrow \infty$ as $t \rightarrow \infty$, then the Lyapunov exponent is $a_* = 0$ and thus SHED enjoys super-linear convergence.
\end{enumerate}
\begin{proof}[Sketch of proof] For 1), we first show that when condition (\ref{eq:condStepSize}) holds, the step size is chosen equal to one by the Armijo backtracking line search. We then show that when also (\ref{eq:condContract}) holds, then the cost converges at least linearly for any subsequent iteration.
 For 2), we upper bound the Lyapunov exponent and exploit local Lipschitz continuity to provide the result. For 3), we exploit at least linear convergence proved in 1) together with the assumption on the cardinality of the set $\mathcal{X}_t$. For the complete proof see Appendix A.
\end{proof}\vspace{-0.3cm}
\end{theorem}
In the above theorem we have shown that SHED enjoys at least linear convergence and provided a sufficient condition on the choice of renewals indices set to guarantee super-linear convergence. The sufficient condition could be easily guaranteed for example with a choice of periodic renewals with a period such that the cardinality of $\mathcal{X}_t$ is big enough. Note that in the case in which all agents send one EEP at each iteration we can get an explicit expression for the Lyapunov exponent also in the convex cost case, getting\vspace{-0.2cm}
\begin{equation}\label{eq:LyapBoundCX}
		a_* \leq \bar{a}_T :=  \prod_{k = 1}^{T}(1 - \frac{\bar{\lambda}_n^{o}}{{\bar{\rho}_{k}^{o}}})^{1/T}.
\end{equation}\vspace{-0.2cm}
\subsection{Heuristics for the choice of $\mathcal{I}$}\label{subsec:heuristicsFib}
\begin{figure}[t!]
\centering
		\includegraphics[width=1\columnwidth]{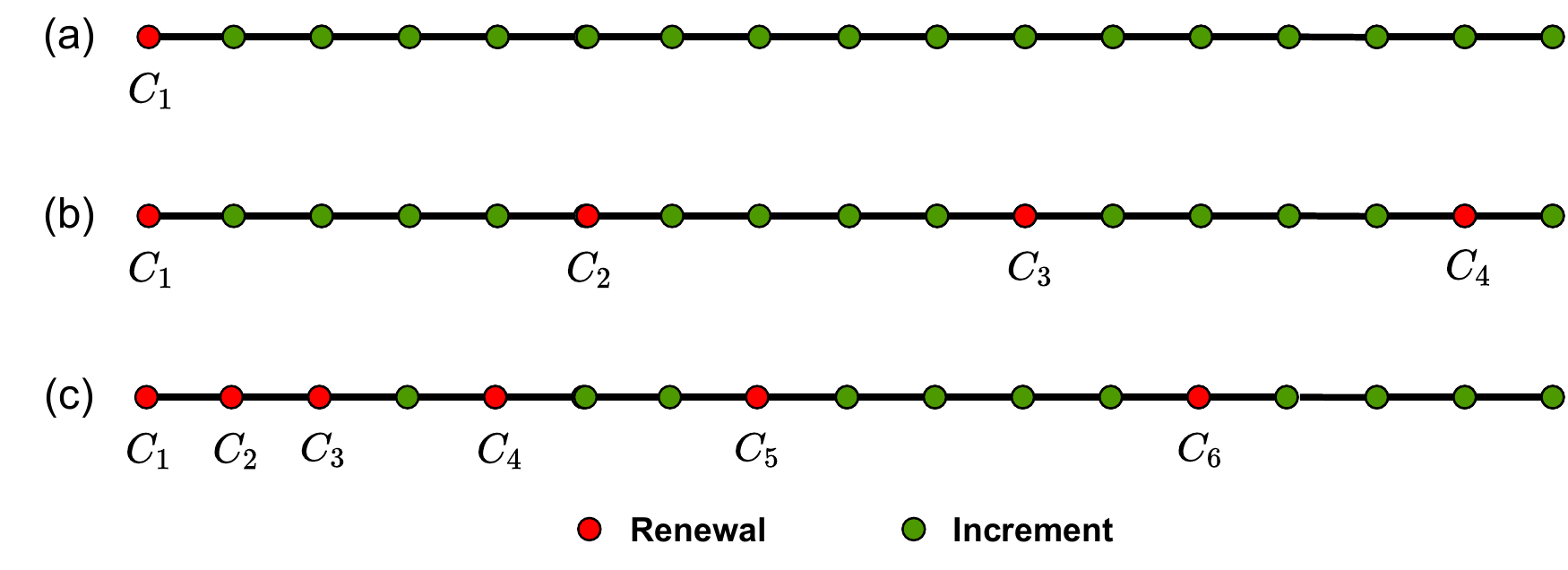}
		\caption{Illustration of possible choices of renewal indices set. The set $\mathcal{I} = \{C_j\}$ specifies the iterations at which a renewal takes place. (a) illustrates the least squares case in which renewal is performed only once, see Algorithm~\ref{alg:1}, (b) the periodic renewals case with $T = 5$, so $\mathcal{I} = \{1, 5, 10, ...\}$, see Algorithm~\ref{alg:variantOf1}, and (c) the set in which the distance between renewals increases according to the Fibonacci sequence.}
		\label{fig:renewalIllustr}  
\end{figure}

\vspace{-0.2cm}
From the theoretical results provided above we can do a heuristic design for the renewal indices set $\mathcal{I}$. In particular, we see from the bound (\ref{eq:convBound}) in Theorem \ref{Th:theorem_2} that when we are at the first iterations of the optimization we would frequently do the renewal operation, given that the Hessian matrix changes much faster and that the term $\|{\x}^{k_t} - {\x}^t\|$ is big. As we converge, instead, we would like to reduce the number of renewal operations, to improve the convergence rate, and this is strongly suggested by the result 2) related to the Lyapunov exponent in Theorem \ref{Th:newLinSupLin}. Furthermore, the super-linear convergence that follows from 3) in the same Theorem suggests to keep performing renewals in order to let the cardinality of $\mathcal{X}_t$ to grow sufficiently fast with $t$.\\
To evaluate SHED performance, we obtain the results of the next section using two different renewal strategies. In the first we choose the distance between renewals to be determined by the \emph{Fibonacci} sequence, so $\mathcal{I} = \{C_j\}$, where $C_j = \sum_{k = 1}^{j}F_k$, $F_k$ being the \emph{Fibonacci} sequence, with $F_0 = 0, F_1 = 1$. When the sequence $C_j$ reaches $n-1$, the next values of the sequence are chosen so that $C_{j+1} = C_j + n-1$. We will call this method Fib-SHED. The second strategy is based on the inspection of the value of the gradient norm, $\|\mathbf{g}_t\|$, which is directly related to $\|{\x}^t - {\x}^*\|$. In particular, we will make a decision concerning renewals at each iteration by evaluating the empirically observed decrease in the gradient norm. If $\|\mathbf{g}_t\| - \|\mathbf{g}_{t-1}\| < b(\|\mathbf{g}_{t-1}\| - \|\mathbf{g}_{t-2}\|)$ for some constant $b$, this strategy triggers a renewal. To guarantee at least linear convergence, we impose a renewal after $n$ iterations in which no renewal has been triggered. In the rest of the paper we call this strategy GN-SHED (Gradient Norm-based SHED). See Fig.~\ref{fig:renewalIllustr} for an illustration of some different possible choices for the renewal indices sets.

%% file: JMLR/section_6.tex
In this section we present empirical results obtained with real datasets. In particular, to illustrate Algorithm \ref{alg:1}, thus in the case of regression via least squares, we consider the popular Million Song (1M Songs) Dataset \cite{1MSongs} and the Online News Popularity dataset \cite{NewsPop}, both taken from the UCI Machine Learning Repository \cite{UCIRep}. For the more general non linear convex case, we apply logistic regression to two image datasets, in particular the FMNIST \cite{FMNIST} and EMNIST digits \cite{EMNIST} datasets, and to the ‘w8a’ web page dataset available from the libSVM library \cite{LIBSVM}. First, we show performance assessments related to the theoretical results of the previous sections. In particular, we show how the versatility of the algorithm applies effectively to the FL framework, showing results related to parameters choice and heterogeneity of CRs. Finally, we compare our algorithm against state-of-the-art approaches in both i.i.d. and non i.i.d. data distributions, showing the advantage that can be provided by our approach, especially in the case of heterogeneous CRs. When comparing with other algorithms, we provide results for both i.i.d. and non i.i.d. partitions to show that our algorithm is competitive with state-of-the-art approaches also in the i.i.d. configuration.

In the general convex case, we follow the heuristics discussed at the end of Section \ref{sec:generalConvex}, where the version using the Fibonacci sequence to define $\mathcal{I}$ is called Fib-SHED while the event-triggered one based on the gradient norm inspection is called GN-SHED. As done previously in the paper, Algorithm \ref{alg:1} is referred to as SHED-LS, and Algorithm \ref{alg:2} as SHED, while their periodic variants (Algorithm \ref{alg:variantOf1}), are referred to as SHED-LS-periodic and SHED-periodic, respectively. If the value of $d_t^{(i)}$ is not specified, SHED-LS and SHED stand for the proposed approach when $d_t^{(i)} = 1$ for each iteration $t$ and agent $i$, $i = 1, ..., M$.

In this section, we show results for $\mu = 10^{-5}, 10^{-6}, 10^{-8}$, and we have obtained similar results with $\mu = 10^{-4}$. Note that $10^{-4}, 10^{-6}, 10^{-8}$ were the values considered in \cite{GIANT}.
\subsection{Datasets}
We consider the following datasets. For each dataset, we partition the dataset for the FL framework in both an i.i.d. and non i.i.d. way. In particular:
\subsubsection{Least squares}\label{subsubsec:synthData} We consider the popular Milion Song Dataset \cite{1MSongs}. The objective of the regression is the prediction of the year in which a song has been recorded. The dataset consists in 515K dated tracks starting from 1922 recorded songs. We take a subset of 300K tracks as training set. The feature size of each data sample is $n = 90$. As preprocessing of the dataset we apply rescaling of the features and labels to have them comprised between $0$ and $1$. To get a number of samples per agent in line with typical federated learning settings, we spread the dataset to $M = 200$ agents, so that each agent has around $800$ data samples.

As a non i.i.d configuration, we consider a scenario with \emph{label distribution skew} and unbalancedness (see \cite{advances}). In particular, we partition the dataset so that each agent has songs from only one year, and different number of data samples. In this configuration, some agent has a very small number of data samples, while other agents have up to 11K data samples.

We also provide some results using the Online News Popularity \cite{NewsPop}, made of $40$K online news data samples. The target of the prediction is the number of shares given processed information about the online news. Data is processed analogously to 1M Songs. $M = 30$ agents are considered with $500$ data sample each.
\subsubsection{Logistic regression}\label{FMNIST} To validate our algorithm with a non linear convex cost we use two popular and widely adopted image datasets and a web page prediction dataset.\\
\textbf{Image datasets:}\ \ We consider the FMNIST (Fashion-MNIST) dataset \cite{FMNIST}, and the EMNIST (Extended-MNIST) digits dataset \cite{EMNIST}. For both, we consider a training set of $60K$ data samples. We apply standard pre-processing to the datasets: we rescale the data to be between $0$ and $1$ and apply PCA \cite{PCA-MNIST}, getting a parameter dimension of $n = 300$. Given that the number of considered data samples is smaller with respect to 1M-Songs, we distribute the dataset to $M = 28$ agents.

We focus on one-vs-all binary classification via logistic regression, which is the basic building block of multinomial logistic regression for multiclass classification. In a one-vs-all setup, one class (that we call target class), needs to be distinguished from all the others. The i.i.d. configuration is obtained by uniformly assigning to each agent samples from the target class and mixed samples of the other classes. When the agents perform one-vs all, we provide them with a balanced set where half of the samples belong the target class, and the other half to the remaining classes, equally mixed. In this way, each agent has around $400$ data samples.

To partition the FMNIST and EMNIST datasets in a non i.i.d. way, we provide each agent with samples from only two classes: one is the target class and the other is one of the other classes. This same approach was considered in \cite{HeterogeneousDiv}. In this section we show results obtained with a target class corresponding to label 'one' of the dataset, but we have seen that equivalent results are obtained choosing one of the other classes as target class. \\
\textbf{Web page dataset:} \ \ We consider the web page category prediction dataset ‘w8a’, available from libSVM \cite{LIBSVM}. The dataset comprises 50K data samples and the objective of the prediction is the distinction between two classes, depending on the website category. The dataset is already pre-processed. Being a binary and strongly unbalanced dataset, we do not consider an i.i.d. configuration, and spread the dataset to the agents similarly to the case of FMNIST, with the difference that some agent only has data samples from one class. The number of data samples per agent is between $700$ and $900$.
\subsection{Federated backtracking}
To tune the step size when there are no guarantees that a step size equal to one decreases the cost, we adopt the same strategy adopted in \cite{GIANT}: an additional communication round takes place in which each agent shares with the master the loss obtained when the parameter is updated via the new descent direction for different values of the step size. In this way, we can apply a distributed version of the popular Armijo backtracking line search (see Algorithm \ref{alg:lineSearch} or \cite{ConvexOptimization}, page 464). When showing the results with respect to communication rounds, we always include also the additional communication round due to backtracking.
\subsection{Heterogeneous channels model}\label{subsec:hetChs}
As previously outlined, our algorithm is suitable for FL frameworks in which agents have heterogeneous transmission resources, that is for instance the case in Federated Edge Learning. As an example we consider the case of wireless channels, that are characterized by strong variability. In particular, in the wireless channel subject to fading conditions and interference, at each communication round agents could have very different CRs available \cite{Pase, ChannelUncertainty, JointWirelessFL}. To show the effectiveness of our algorithm in such a scenario, we consider the Rayleigh fading model. In a scenario with perfect channel knowledge, in which agent $i$ allocates a certain bandwidth $B^{(i)} = B$ (we consider it to be the same for all agents) to the FL learning task, we have that the achievable rate of transmission for the $i$-th user is:
\begin{equation}
	R^{(i)} = B\log_2(1 + \gamma\Gamma^{(i)})
\end{equation}
where $\Gamma^{(i)}$ is a value related to transmission power and environmental attenuation for user $i$. For simplicity we fix $\Gamma^{(i)} = \Gamma = 5$ for all users (in \cite{Pase}, for instance, $\Gamma = 1$ and $\Gamma = 10$ were considered). The only source of variability is then $\gamma \sim \mathit{Exp}(\nu)$, modeling the Rayleigh fading effect. We fix $\nu = 1$. 

With respect to our algorithm, for illustration purposes, we compare scenarios in which the number of vectors in $\mathbb{R}^{n}$ related to second order information, $d_t^{(i)}$, is deterministically chosen and fixed for all users, with a scenario in which $d_t^{(i)}$ is randomly chosen for each round for each user. In particular, we fix:
\begin{equation}\label{eq:defOfdGamma}
	d_t^{(i)} = d_{\gamma} := \floor{d_0\log_2(1 + \gamma\Gamma)}.    
\end{equation}
This would correspond to each user allocating a bandwidth of $B = d_0B_0$ to the transmission of second order information, where $B_0$ is the transmission speed needed to transmit a vector in $\mathbb{R}^{n}$ per channel use. In the rest of the paper, we fix $d_0 = 2$. This implies that in average the number of vectors that an agent is able to transmit is $4$, and the actual number can vary from $0$ (only the gradient is sent) to $10$. For simplicity, when adopting this framework, we assume that agents are always able to send at least a vector in $\mathbb{R}^n$ per communication round (so the gradient is always sent). Further simulations are left as future work.

\subsection{Choice of $\rho_t$}
\begin{figure}[t!]
	\centering
	\includegraphics[width=0.9\columnwidth]{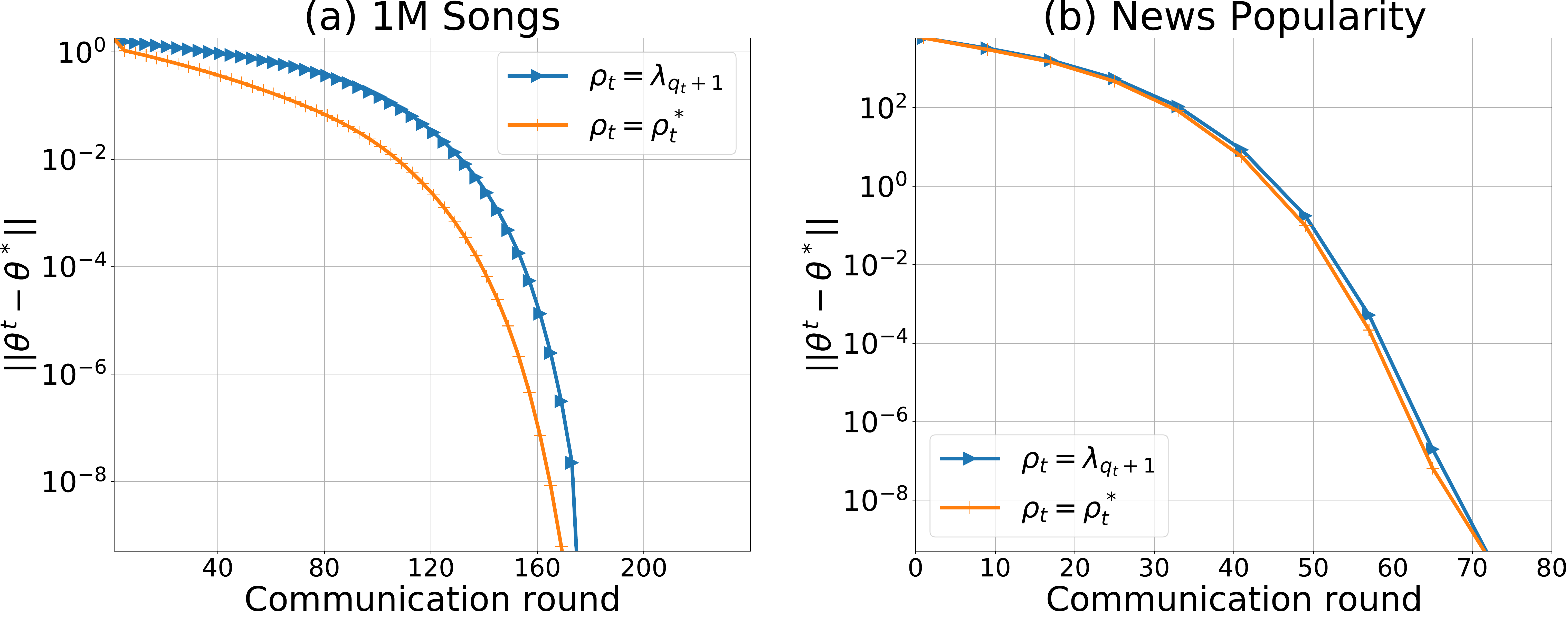}
	\caption{Performance comparison for different values of $\rho_t$. The best choice in terms of estimation error from Corollary \ref{th:corollaryEE} is compared against the choice that was proposed in \cite{MontanariWay}, that is $\rho_t = \lambda_{q_t+1}$}
	\label{fig:rhoComp}
\end{figure}

In Fig. \ref{fig:rhoComp} we analyze the optimization performance with respect to the choice of the approximation parameter $\rho_t$, following the theoretical results of Section \ref{sec:centralLSs}, in particular Theorem \ref{lemma_1} and Corollary \ref{th:corollaryEE}. We provide performance analysis in the distributed setting for least squares with the 1M Song and Online News Popularity datasets in (a) and (b), respectively. We show the error $\|\theta^t - \theta^*\|$ versus the number of communication rounds. We adopt the notation of Section \ref{sec:generalConvex} as it includes least squares as a special case. In this subsection, the same parameter choice is done for all the agents, thus we omit to specify that the parameter is specific of the $i$-th agent. We compare two choices for the approximation parameter: (i) the one providing the best estimation error according to Corollary \ref{th:corollaryEE}, $\rho_t^* = (\hat{\lambda}_{q_t+1, t} + \hat{\lambda}_{n, t})/2$ with the step size $\eta_t = 1$, and (ii) the one that was proposed in \cite{MontanariWay}, where $\rho_t = \hat{\lambda}_{q_t+1, t}$. In the latter case, following the result of Theorem \ref{lemma_1} related to the best convergence rate we pick the step size to be equal to the $\eta_t^*$ in Eq. (\ref{eq:bestEta}). From the plots, we can see the improvement provided by the choice of $\rho_t^*$ against $\hat{\lambda}_{q_t+1, t}$, although the performance is very similar for both values of the approximation parameter.
\subsection{Lyapunov exponent convergence bound}
\begin{figure}[t!]
	\centering
	\includegraphics[width=0.9\columnwidth]{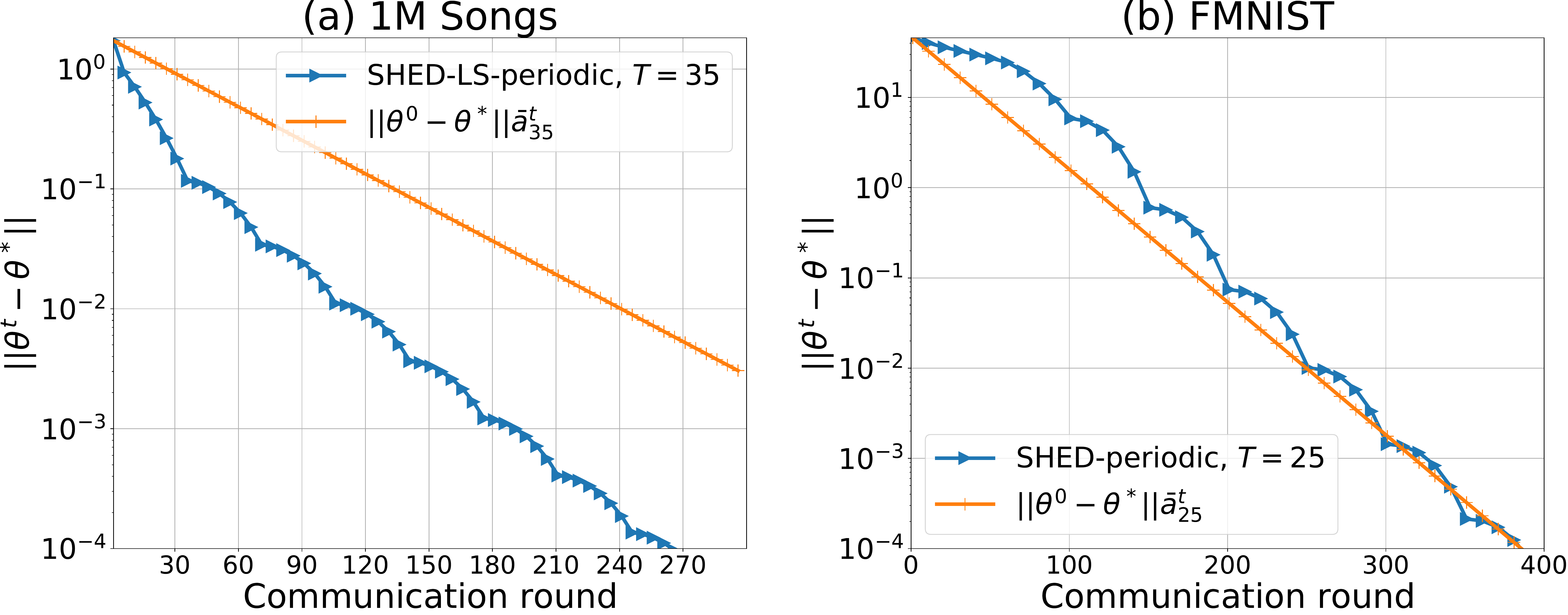}
	\caption{Linear convergence with periodic renewals illustrated via the study of the upper bound on the dominant Lyapunov exponent of the estimation error from equation (\ref{eq:LyapExp}) and point 2) of Theorem \ref{Th:newLinSupLin}. $\Bar{a}_T$ is as in eqs. (\ref{eq:LyapExp}) and (\ref{eq:LyapBoundCX}) for (a) and (b), respectively. Note that the upper bound is on the slope of the decreasing cost.} 
	\label{fig:Lyap}  
\end{figure}

In Fig. \ref{fig:Lyap}, we illustrate how the Lyapunov exponent bounds derived in Sec. \ref{subsec:Periodic} and Theorem \ref{Th:newLinSupLin}-2) characterize the linear convergence rate of the algorithms. In particular, we consider the cases of periodic renewals in both the least squares and logistic regression case, considering the case in which $q_t^{(i)} = q_{t-1}^{(i)} +1$, $\forall i, t$ and renewals are periodic with period $T$. The plots show how the linear convergence rate is dominated by the Lyapunov exponent bound characterized by $\Bar{a}_T$. For illustrative purposes, we show the results for the choice of $T = 35$ and of $T = 25$ for least squares on 1M songs and logistic regression on FMNIST, respectively.
\subsection{Role of $d_t$}
\begin{figure}[t!]
	\centering
	\includegraphics[width=0.9\columnwidth]{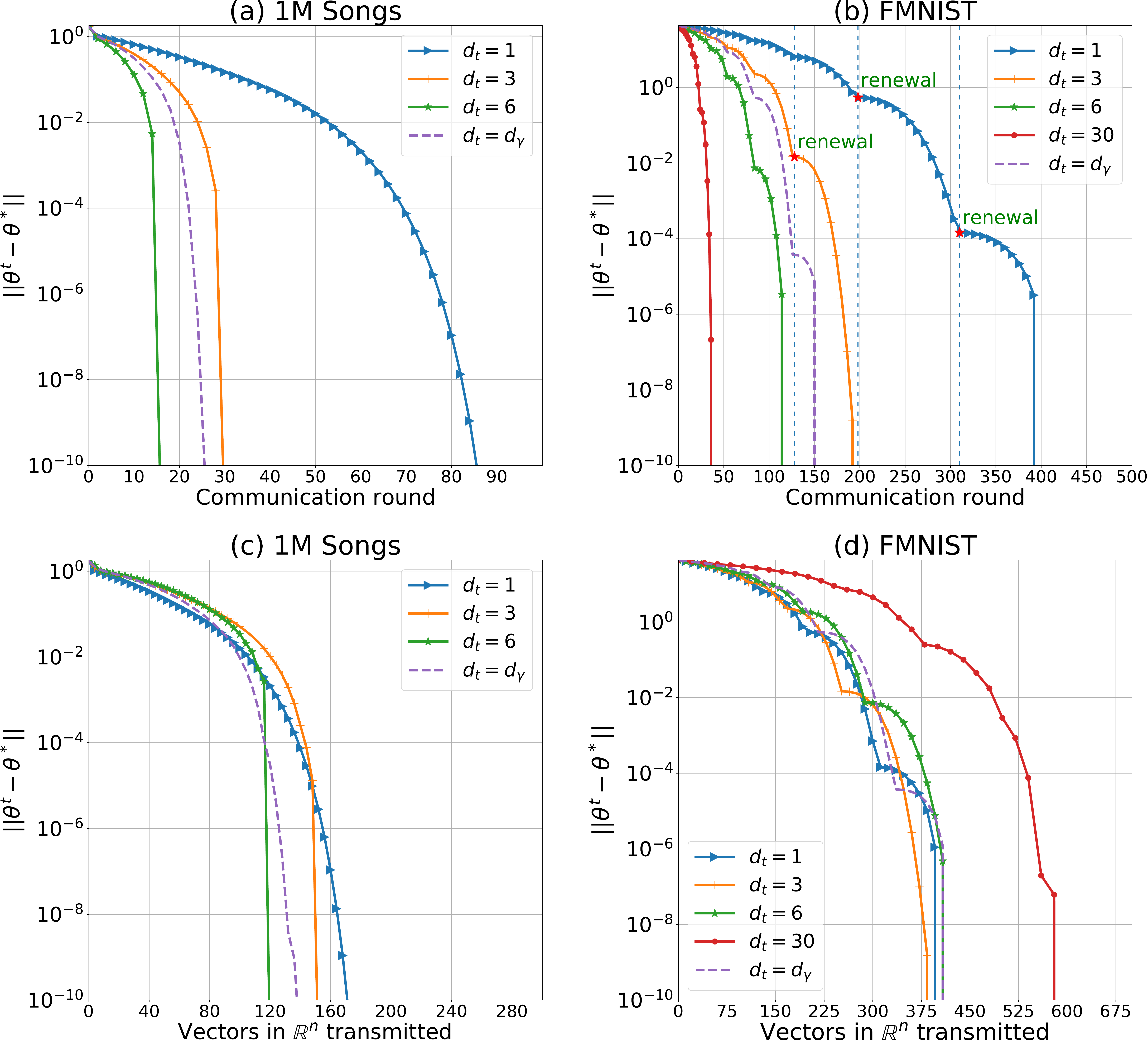}
	\caption{Performance comparison for different values of $d_t$ for (i) linear regression on 1M Songs ((a) and (c)) and (ii) logistic regression on FMNIST ((b) and (d)). $d_{\gamma}$ is as defined in Eq. (\ref{eq:defOfdGamma}), with $d_0 = 2$, so $d_{\gamma}$ is in average equal to $4$. In (b), we emphasize three points where the renewal operation (see Algorithm \ref{alg:2}) takes place.}
	\label{fig:IncrComp}  
\end{figure}

In Figure \ref{fig:IncrComp}, we show the impact of each agent transmitting more eigenvalue-eigenvector pairs per communication rounds, showing the optimization results when the increment, $d_t^{(i)}$, takes different values. In this subsection, for illustration purposes, renewals of SHED are determined by the Fibonacci sequence (see Sec. \ref{subsec:heuristicsFib}). We consider the case when the number is the same and fixed (specifically we consider $d_t^{(i)} \in \{1, 3, 6, 30\}$) for all the agents and the case $d_t^{(i)} = d_{\gamma}$, where $d_{\gamma}$ is as defined in Eq. (\ref{eq:defOfdGamma}), with $d_0 = 2$ and $\Gamma = 5$, that means that in average $d_{\gamma}$ is equal to $4$. In this latter case each agent is able to transmit a different random number of eigenvalue-eigenvector pairs. This configuration is relevant as our algorithm allows agents to contribute to the optimization according to their specific communication resources (CRs). In Figures \ref{fig:IncrComp}-(a)-(c) we show the results for the least squares on 1M Songs, while in Figures \ref{fig:IncrComp}-(b)-(d) we show the results in the convex case of logistic regression on the FMNIST dataset. We can see from Figures \ref{fig:IncrComp}-(a)-(b) how the global number of communication rounds needed for convergence can be significantly reduced by increasing the amount of information transmitted at each round. In particular, at each round, the number of vectors in $\mathbb{R}^n$ being transmitted is $d_t + 1$, since together with the $d_t$ scaled eigenvectors, $\{\Tilde{\mathbf{v}}_{j, t}^{(i)}\}_{j = q^{(i)}_{t-1}+1}^{q^{(i)}_t}$ (see Algorithm 2), agents need to transmit also the gradient. 
In the least squares case, the number of iterations needed for convergence is surely smaller than the number of EEPs sent: when the $n-1$-th EEP has been shared, convergence surely occurs. When the number of EEPs is random ($d_{\gamma}$), but equal to $4$ in average, we see that we can still get a significant improvement, which is between the choices $d_t = 3$ and $d_t = 6$. 

Let us now analyze the case of logistic regression. In Figures \ref{fig:IncrComp}-(b)-(d) we emphasize the role of the \emph{renewal} operation, showing also how incrementally adding eigenvectors of the outdated Hessian improves the convergence, as formalized and shown in Theorems \ref{Th:theorem_2} and \ref{Th:newLinSupLin}. In particular, from Figure \ref{fig:IncrComp}-(b) it is possible to appreciate the impact of increasing the interval between renewals.

Quantifying the improvement provided by the usage of greater increments analyzing the empirical results, we see that when $d_t = 3$, the number of vectors transmitted per round is $4$, against the $2$ transmitted when $d_t = 1$, so the communication load per round is twice as much. We see that, despite this increase in data transmitted per round, in the case $d_t = 3$ the overall number of communication rounds is halved with respect to the case $d_t = 1$. An equivalent result occurs for the case $d_t = 6$. Notably, for the case $d_t = d_{\gamma}$, with an average increment equal to $4$, we see that the convergence speed is in between the cases $d_t = 3$ and $d_t = 6$, which shows the effectiveness of the algorithm under heterogeneous channels. For $d_t = 30$, even if we increase the communication by a factor $15$, we get a convergence that is only $10$ times faster. This is of interest because it shows that if we increase arbitrarily the increment $d_t$, we pay in terms of overall communication load. 

In Figures \ref{fig:IncrComp}-(c)-(d) we plot the error as a function of the amount of data transmitted per agent, where the number of vectors in $\mathbb{R}^n$ is the unit of measure. These plots show that, for small values of $d_t$ (in particular, $d_t \in \{1, 3, 6\}$), the overall data transmitted does not increase for the considered values of $d_t$, meaning that we can significantly reduce the number of global communication rounds by transmitting more data per round without increasing the overall communication load. This is true in particular also for the case $d_t = d_{\gamma}$, thus when the agents' channels availability is heterogeneous at each round, showing that our algorithm works even in this relevant scenario without increasing the communication load. On the other hand, in the case of $d_t = 30$, if we increase the amount of information by an order of magnitude, even if we get to faster convergence, we have as a consequence a significant increase in the communication load that the network has to take care of.

\subsection{Comparison against other algorithms}\label{subsec:otherAlgs}
In this subsection, SHED-LS+ and SHED+ signifies that the number of Hessian EEPs is chosen randomly for each agent $i$, according to the model for simulating heterogeneous CRs illustrated in Sec. \ref{subsec:hetChs}, so $d_t^{(i)} = d_{\gamma}$, with an average increment equal to $4$. To distinguish the different heuristic strategies for the renewals indices set, we use Fib-SHED to denote the approaches where $\mathcal{I}$ is pre-determined by the Fibonacci sequence, while GN-SHED to denote the approach in which renewals are event-triggered by the gradient norm inspection (see Sec. \ref{subsec:heuristicsFib}) 

We compare the performance of our algorithm with a state-of-the art first-order method, accelerated gradient descent (AGD, with the same implementation of \cite{GIANT}). 

As benchmark second-order methods we consider the following:
\begin{itemize}
	\item a distributed version of the Newton-type method proposed in \cite{MontanariWay}, to which we refer as Mont-Dec, which is the same as Algorithm \ref{alg:2} with the difference that the renewal occurs at each communication round, so the Hessian is always recomputed and the second-order information is never outdated. In this way, the eigenvectors are always the first ($\{{v_{j,t}}\}_{j = 1}^{d_t}$), so this algorithm is neither incremental nor exploits outdated second-order information. We fix the amount of second-order information sent by the agents to be the same as the previously described SHED+.
	\item the distributed optimization state-of-the-art NT approach exploiting the harmonic mean, GIANT \cite{GIANT}. While in GIANT the local Newton direction is computed via conjugate gradient method, in DONE \cite{DONE}, a similar approach has been proposed but using Richardson iterations. To get our results, we provide the compared algorithm with the actual exact harmonic mean, so that we include both the algorithms at their best. Referring to this approach, we write GIANT in the figures. We remark that GIANT requires an extra communication round, because the algorithm requires the agents to get the global gradient. We also implemented the determinantal averaging approach \cite{DetAvg} to compensate for the inversion bias of the harmonic mean, but we did not see notable improvements so we do not show its results for the sake of plot readability.
	\item the very recently proposed FedNL method presented in \cite{FedNL}. We adopted the best performing variant (FedNL-LS), with rank-1 matrix compression, which initializes the Hessian at the master with the complete Hessian computed at the initial parameter value, so the transmission of the Hessian matrix $\mathbf{H}(\x^0)$ at the first round is required. This method requires the computation of the Hessian and of an $n\times n$ matrix SVD at each iteration. FedNL uses the exact same amount of CRs as SHED (so with $d_t^{(i)} = 1, \forall i$), as it sends the gradient and one eigenvector at each iteration. For the Hessian learning step size, we adopt the same choice of \cite{FedNL} (step size equal to 1), and we observed that different values of this parameter did not provide improvements.
\end{itemize} 
We do not compare Mont-Dec and FedNL in the least squares case, as these algorithms are specifically designed for a time-varying Hessian and it does not make sense to use them in a least squares problem.\\
\begin{figure}[b!]
	\centering
	\includegraphics[width=0.85\columnwidth]{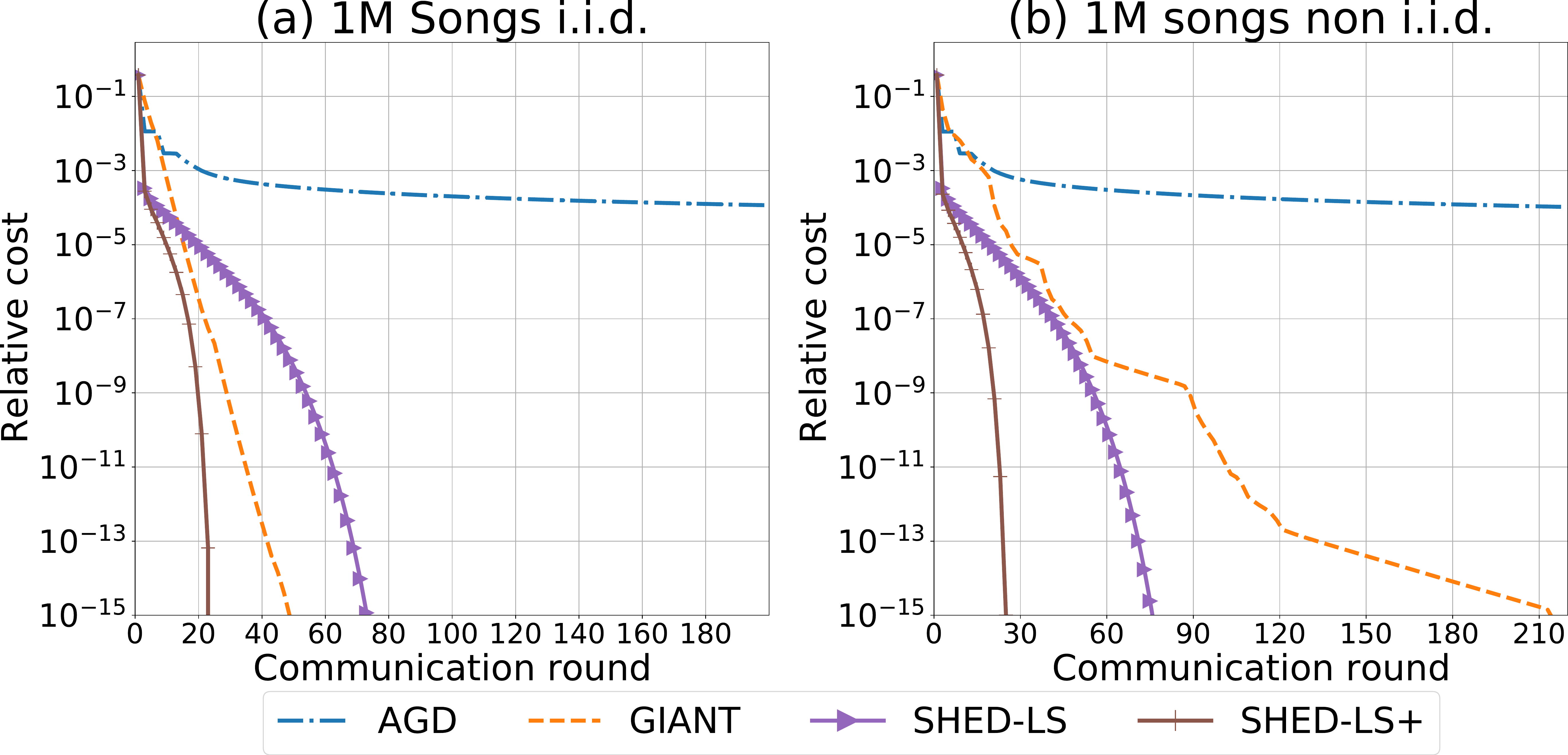}
	\caption{Performance comparison for least squares on 1M Songs. Relative cost is $f({\x}^t) - f({\x}^*)$. For SHED-LS and SHED-LS+, see Algorithm \ref{alg:1}}
	\label{fig:LS_Comp}
\end{figure}
In the following, we show the details of results for the least squares on 1M Songs dataset and then for logistic regression on the FMNIST dataset. The detailed Figures of convergence results obtained with the EMNIST digits and ‘w8a’ are provided in Appendix B, while a summary of the results obtained on all the considered datasets is provided in Section \ref{subsec:summary}.
\subsubsection{Least squares on 1M Songs} 
In Fig. \ref{fig:LS_Comp} we show the performance of the algorithm in the case of least squares applied on the 1M Songs dataset (thus the procedure is the one described in Algorithm 1). We plot the relative cost $f(\theta^t) - f(\theta^*)$ versus the communication rounds.  We see that AGD is characterized by very slow convergence due to the high condition number. Together with AGD, in Fig. \ref{fig:LS_Comp} we show three Newton-type approaches: the state-of-the-art GIANT \cite{GIANT}, SHED and SHED+. We see how in the i.i.d. case GIANT performs well even if SHED+ is the fastest, but all the three algorithms have similar performance compared to the first-order approach (AGD). In the non i.i.d. case, whose framework has been described in Sec. \ref{subsubsec:synthData}, we can see how GIANT has a major performance degradation, even if still obtaining a better performance with respect to first-order methods. On the other hand, both SHED and SHED+ do not show performance degradation in the non i.i.d. case.
\subsubsection{Logistic regression on FMNIST}
\begin{figure}[t!]
	\centering
	\includegraphics[width=0.85\columnwidth]{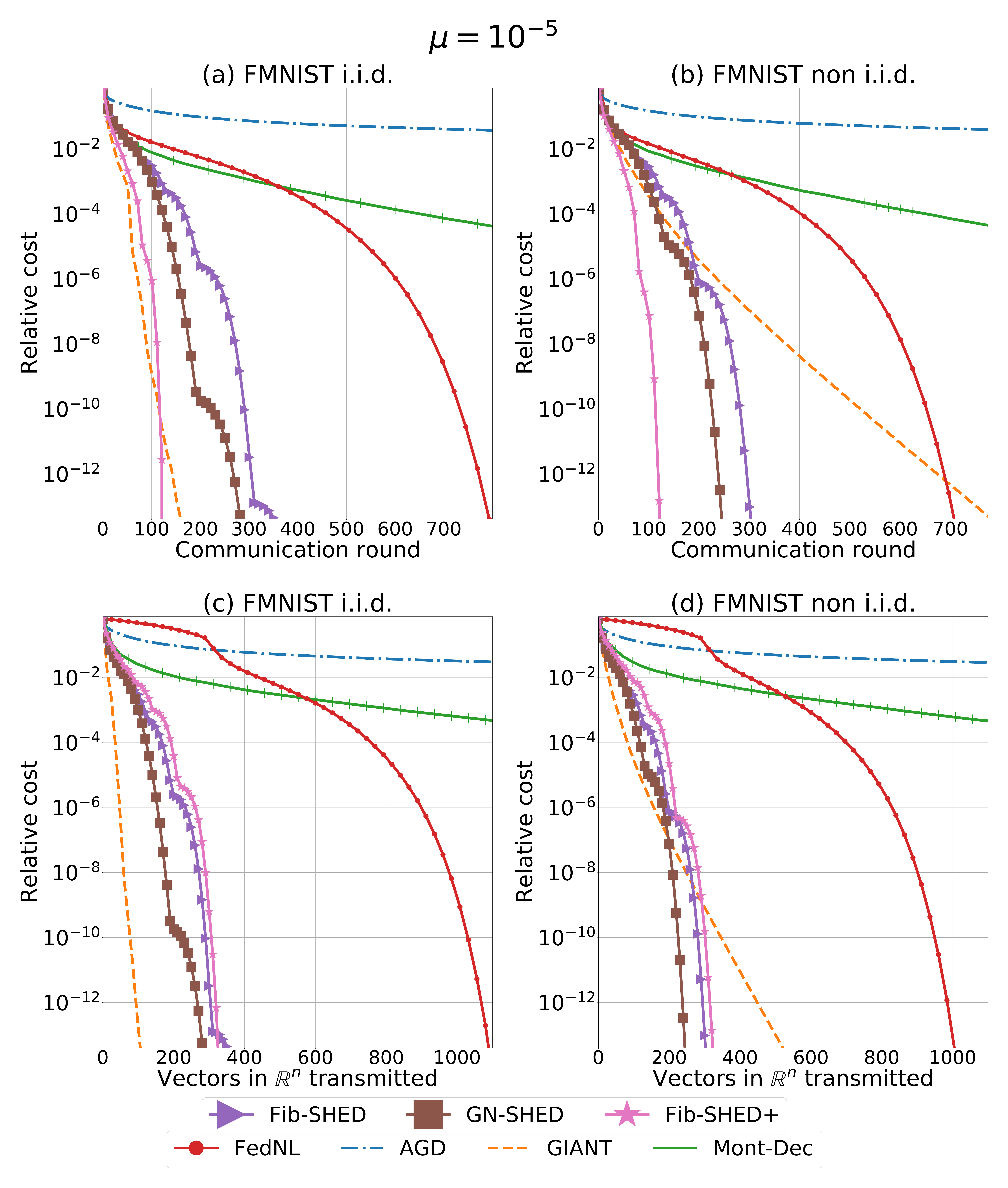}
	\caption{Performance comparison of logistic regression on FMNIST when $\mu = 10^{-5}$. Relative cost is $f({\x}^t) - f({\x}^*)$.}
	\label{fig:FMNIST_Comp}
\end{figure}
\begin{figure}[t!]
	\centering
	\includegraphics[width=0.85\columnwidth]{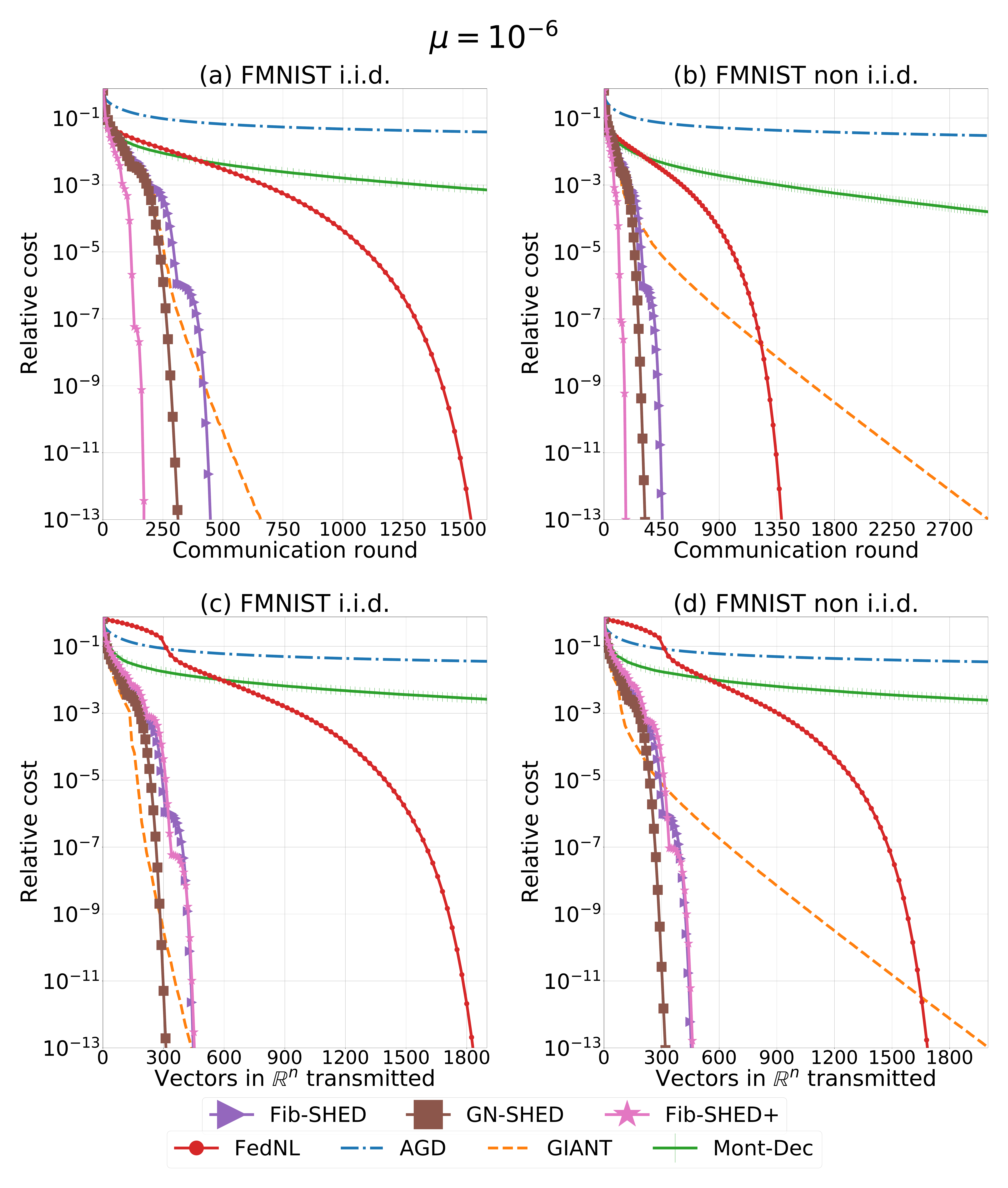}
	\caption{Performance comparison of logistic regression on FMNIST when $\mu = 10^{-6}$. Relative cost is $f({\x}^t) - f({\x}^*)$.}
	\label{fig:FMNIST_CompLowerMu}
\end{figure}

In Figs. \ref{fig:FMNIST_Comp}, \ref{fig:FMNIST_CompLowerMu} we show the results obtained with the FMNIST dataset with the setup described in Sec. \ref{FMNIST}. We show the performance for two values of the regularization parameter, $\mu$, in particular $\mu = 10^{-5}$ and $\mu = 10^{-6}$. We compare our approach (Algorithm \ref{alg:2}, SHED) also against FedNL and Mont-Dec. 
We show the performance of Fib-SHED, Fib-SHED+ and GN-SHED. As we did for Fig. \ref{fig:IncrComp}, to provide a complete comparison, in Figs. \ref{fig:FMNIST_Comp}-\ref{fig:FMNIST_CompLowerMu}-(c)-(d) we show also the relative cost versus the overall amount of data transmitted, in terms of overall number of vectors in $\mathbb{R}^n$ transmitted.
From Fig. \ref{fig:FMNIST_Comp}-(a)-(b), we can see how also in the FMNIST case the non i.i.d. configuration causes a performance degradation for GIANT, while SHED and FedNL are not impacted.

In the i.i.d. case, for $\mu = 10^{-5}$ (Fig. \ref{fig:FMNIST_Comp}), we see that GN-SHED, Fib-SHED, Fib-SHED+ and GIANT require a similar number of communication rounds to converge, while the amount of overall data transmitted per agent is smaller for GIANT. When $\mu = 10^{-6}$, instead, we see from Figure \ref{fig:FMNIST_CompLowerMu} that GIANT requires more communication rounds to converge while the same amount of information needs to be transmitted. Hence, in this case GIANT is more impacted by a lower condition number when compared to our approach. On the other hand, FedNL in both cases shows a much slower convergence speed with respect to Fib-SHED and GN-SHED (approaches requiring the same per-iteration communication load as FedNL). Notice, for FedNL, in Figures \ref{fig:FMNIST_Comp}-(c)-(d) and \ref{fig:FMNIST_CompLowerMu}-(c)-(d), the impact that the transmission of the full Hessian matrices at the
first round has on the overall communication load.

In the considered non i.i.d. case, the large advantage that our approach can provide with respect to the other considered algorithms is strongly evident in both data transmitted and communication rounds.

Comparing the SHED approaches against Mont-Dec we see the key role that the \emph{incremental} strategy exploiting \emph{outdated} second order information has on the convergence speed of our approach. Indeed, even though in the first iterations the usage of the current Hessian information provides the same performance of the SHED methods, the performance becomes largely inferior in the following rounds. \\
From a computational point of view, both the Mont-Dec and the FedNL approach are much heavier than SHED as they require that each agent recomputes the Hessian and SVD at each round. SHED, instead, in the more challenging case of $\mu = 10^{-6}$, requires the agents to compute the Hessian matrix only 12 times out of the 450 rounds needed for convergence. For more details on this, see Figure \ref{fig:ComputIOSVFED}. 

\begin{figure}[t!]
	\centering
	\includegraphics[width=0.85\columnwidth]{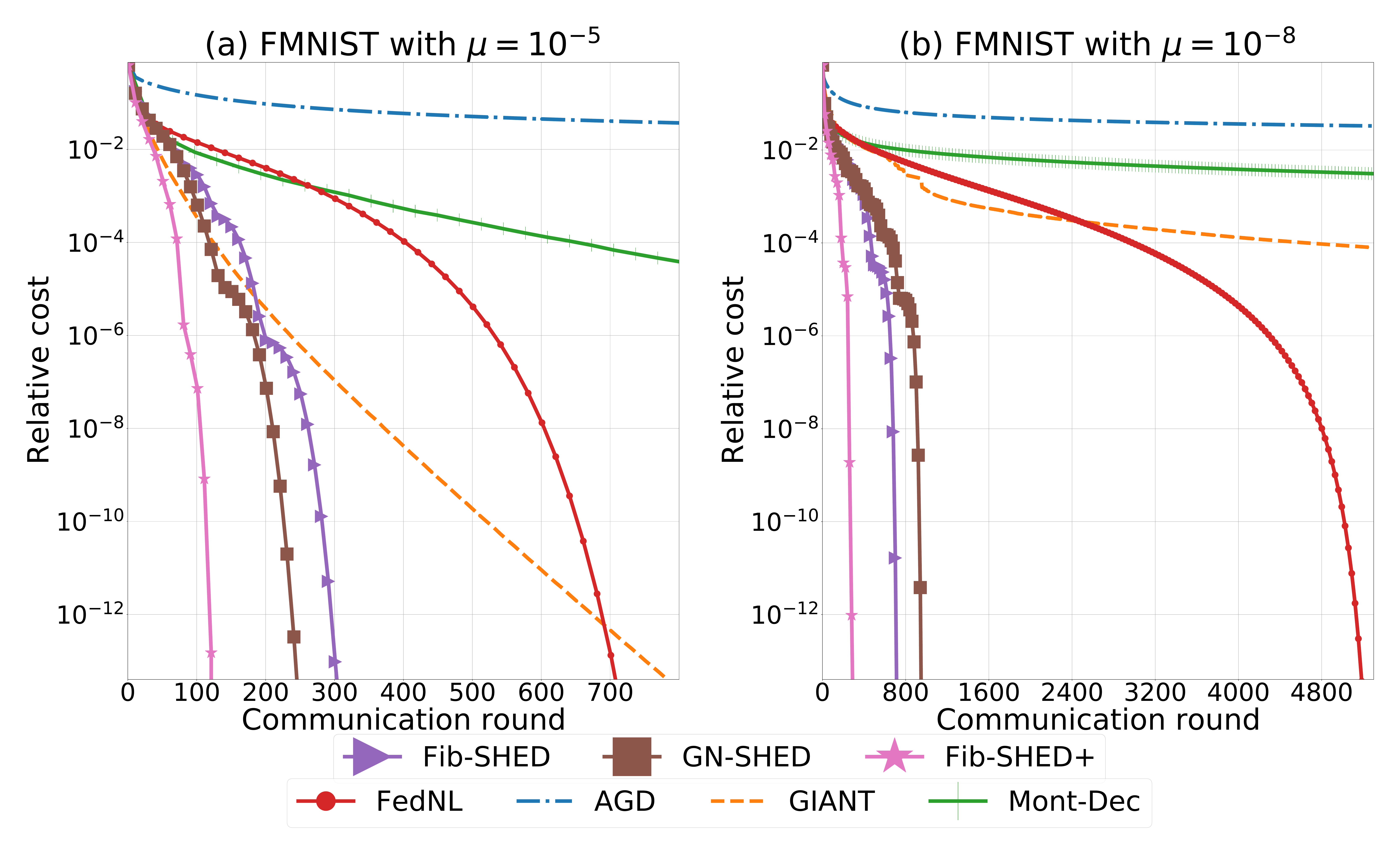}
	\caption{Performance of logistic regression on FMNIST, comparing $\mu = 10^{-5}$ and $\mu = 10^{-8}$. Relative cost is $f({\x}^t) - f({\x}^*)$.}
	\label{fig:FMNIST_CompIllCond}
\end{figure}

In Fig. \ref{fig:FMNIST_CompIllCond}, we show how SHED is much more resilient to ill-conditioning with respect to competing algorithms, by comparing the convergence performance when the regularization parameter is $\mu = 10^{-5}$ and $\mu = 10^{-8}$. With respect to FedNL, note how Fib-SHED worsens its performance with $\mu = 10^{-8}$ by being around 2.5 times slower compared to $\mu = 10^{-5}$, while FedNL worsens much more, being more than 8 times slower with $10^{-8}$ compared to $\mu = 10^{-5}$.

\subsubsection{Comparison on computational complexity}\label{subsec:summary}
\begin{figure}[t!]
	\centering
	\includegraphics[width=1.1\columnwidth]{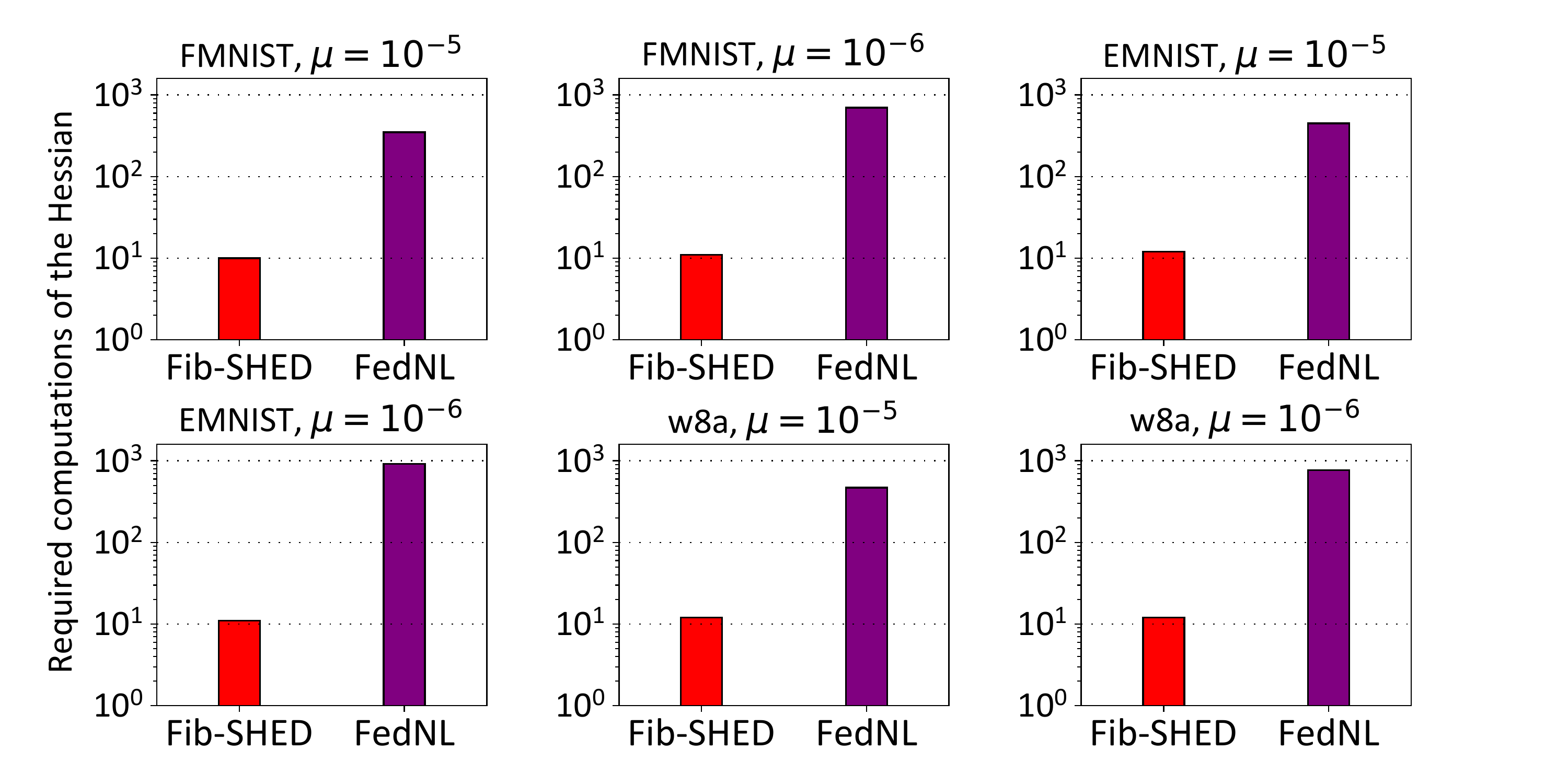}
	\caption{In this plots, we show, for three datasets (FMNIST, EMNIST and w8a), the number of times an agent is required to compute the local Hessian matrix in order for an algorithm to converge, comparing the proposed Fib-SHED and the FedNL \cite{FedNL} algorithms.}
	\label{fig:ComputIOSVFED}
\end{figure}
In Appendix B, we show the results obtained with the EMNIST and w8a datasets. We omit the comparison with GN-SHED given that the results are similar to the ones obtained with Fib-SHED. In the case of EMNIST, we obtain results similar to FMNIST, except that, when $\mu = 10^{-5}$, GIANT is not much impacted by the considered non i.i.d. configuration. Even if in that case GIANT seems to be the best choice, all the other results show that GIANT and related approaches based on the harmonic mean (like DONE) are strongly sensitive to non i.i.d. data distributions. With image datasets (EMNIST and FMNIST), FedNL is largely outperformed by our approach (by both Fib-SHED and Fib-SHED+), while, with the ‘w8a’ dataset, FedNL is more competitive. However, the Fib-SHED and Fib-SHED+ approaches have the very appealing feature that they require agents to compute the local Hessian matrices only \emph{sporadically}. The FedNL approach, instead, requires that the Hessian is recomputed by agents at each round, implying a much heavier computational demand. To better illustrate and quantify this advantage, we show, in Figure \ref{fig:ComputIOSVFED}, the number of times that an agent is required to compute the local Hessian matrix in order to obtain convergence, comparing Fib-SHED and FedNL, in the cases of the three datasets. In the case of EMNIST and FMNIST, we are showing the non i.i.d. configurations, but similar results can be obtained with the i.i.d. ones. The results show that, compared with Fib-SHED, the number of times agents are required to compute the Hessian is always at least ten times greater for FedNL to converge.

%% file: JMLR/Appendix_A.tex
\section*{Proof of Theorem \ref{corollary:convergence}}
First, we need to prove the following Lemma:
\begin{lemma}\label{lemma:2}
	If $\|\mathbf{g}_{t}\| > \omega > 0$, for $\hat{\mathbf{H}}_t$ defined in (\ref{eq:aggregation}), $\mathbf{p}_t = \hat{\mathbf{H}}_t^{-1}\mathbf{g}_t$ there are $\gamma_t, \eta_t > 0$ such that
	\begin{equation}\label{eq:costDecrease}
		f(\boldsymbol{\theta}^{t} - \eta_t \mathbf{p}_t) \leq f(\boldsymbol{\theta}^t) - \gamma_t,
	\end{equation}
	in particular, for a backtracking line search with parameters $\alpha \in (0, 0.5), \beta \in (0, 1)$, it holds:
	\begin{equation}\label{eq:costDecr}
		\gamma_t = \alpha\beta\frac{\Bar{\rho}_t}{K^2}\omega^2
	\end{equation}
	\begin{proof}
		The proof is the same as the one provided in \cite{ConvexOptimization} for the damped Newton phase (page 489-490), with the difference that the "Newton decrement", that we denote by $\sigma_t$, here is $\sigma_t^2 := \mathbf{g}_t^T\mathbf{p}_t = \mathbf{g}_t^T\hat{\mathbf{H}}_t^{-1}\mathbf{g}_t = \mathbf{p}_t^T\hat{\mathbf{H}}_t\mathbf{p}_t$. Furthermore, the property $\hat{\mathbf{H}}_t \geq \Bar{\rho}_t\mathbf{I}, \ \forall{t}$ is used in place of strong convexity.
\end{proof}
\end{lemma}
Note that, because of Assumption \ref{ass:localLCont}, it always holds that $\rho_t^{(i)} > 0, \ \ \forall{i}, \forall{t}$, and this implies $\bar{\rho}_t > 0, \forall{t}$. When $\bar{\rho}_t > 0, \forall{t}$, Lemma \ref{lemma:2} implies $\|\mathbf{g}_t\| \rightarrow 0$. Indeed, if not, there would be some $\epsilon > 0$ such that $\|\mathbf{g}_t\| > \epsilon, \ \forall{t}$. But then (\ref{eq:costDecrease}) immediately implies that $f(\boldsymbol{\theta}^t) \rightarrow -\infty$, that contradicts the strong convexity hypothesis. By strong convexity and differentiability of $f$, $g(\boldsymbol{\theta}) = \nabla f(\boldsymbol{\theta}) = 0 \implies \boldsymbol{\theta} = \boldsymbol{\theta}^*$.

\section*{Proof of Theorem \ref{Th:newLinSupLin}}
1)	We first need the following Lemma:
\begin{lemma}\label{Th:stepSizeConvex}
	Let $\bar{\kappa} = \sum_{i = 1}^{M}\kappa_i$, with $\kappa_i$ the strong convexity constant of the cost $f^{(i)}$ of agent $i$, let $K$ be the smoothness constant of $f$ and $M(t) = \max\{\|{\x}^t - {\x}^*\|, \|{\x}^{k_t} - {\x}^*\|\}$. Applying Algorithm~\ref{alg:2}, if	
	\begin{equation}\label{eq:condForEta1}
		3\bar{\kappa}(M(t) + \|{\x}^t - {\x}^*\|) + K\|{\x}^t - {\x}^*\| \leq \frac{3\bar{\kappa}^2}{L}(1-2\alpha),
	\end{equation}
	then, for any $\alpha \in (0, 1/2)$, the backtracking algorithm (\ref{alg:lineSearch}) chooses $\eta_t = 1$. 
	\begin{proof}[Sketch of proof] The proof uses a procedure similar to the one used in \cite{ConvexOptimization}, page 490-491 to prove the beginning of the quadratically convergent phase in the centralized Newton method. Then, leveraging Lipschitz continuity and the fact that, thanks to the choice $\rho_t^{(i)} = \lambda_{q_t^{(i)}}^{(i)}$, it is $\hat{\mathbf{H}}_t \geq \mathbf{H}({\x}^{k_t})$, the result is derived. For the complete proof, see the Proof of Lemma 12 at the end of this Appendix.
	\end{proof}
\end{lemma}	
Let condition (\ref{eq:condForEta1}) be satisfied. Then, if \vspace{-0.3cm}
\begin{equation}\label{eq:condContraction}\vspace{-0.3cm}
	(3/2)L\|{\x}^t - {\x}^*\| + LM(t) \leq \bar{\kappa},
\end{equation}
the convergence of SHED is at least linear. Indeed, if (\ref{eq:condForEta1}) is satisfied, then, from Lemma \ref{Th:stepSizeConvex}, the step size is $\eta_t = 1$ and the convergence bound (see Theorem \ref{Th:theorem_2}) becomes $	\|\boldsymbol{\theta}^{t+1} - \boldsymbol{\theta}^*\| \leq c_t\|\boldsymbol{\theta}^t - \boldsymbol{\theta}^*\|$, with\vspace{-0.2cm}
\begin{equation}\label{eq:lineConvBound}\vspace{-0.2cm}
	\begin{aligned}
		c_t &= (1 - \frac{\bar\lambda_{n, t}}{\Bar{\rho}_t} + \frac{L}{\Bar{\rho}_t}\|\boldsymbol{\theta}^t - \boldsymbol{\theta}^{k_t}\| + \frac{L}{2\Bar{\rho}_t}\|\boldsymbol{\theta}^t - \boldsymbol{\theta}^{*}\|) \\&\leq (1 - \frac{\bar\lambda_{n, t}}{\Bar{\rho}_t} + \frac{L}{\Bar{\rho}_t}\|\boldsymbol{\theta}^{k_t} - \boldsymbol{\theta}^{*}\| + \frac{3L}{2\Bar{\rho}_t}\|\boldsymbol{\theta}^t - \boldsymbol{\theta}^{*}\|)
	\end{aligned}
\end{equation}
and it is easy to see that condition (\ref{eq:condContraction}) implies that $c_t < 1$ and thus we get a contraction in $\|{\x}^t - {\x}^*\|$. Furthermore, when conditions (\ref{eq:condForEta1}) and (\ref{eq:condContraction}) are both satisfied at some iteration $\bar{t}$, they are then satisfied for all $t \geq \bar{t}$ and thus $c_t < 1$ for all $t \geq \bar{t}$. Indeed, $c_{t} < 1$ implies that $\|{\x}^{t+1} - {\x}^*\| < \|{\x}^t - {\x}^*\|$ and $M(t+1) \leq M(t)$ because either $k_{t+1} = k_t$ or $k_{t+1} = t+1$. Note that Assumption \ref{th:AssDelayBound} is needed to guarantee that (\ref{eq:condForEta1}) and (\ref{eq:condContraction}) are eventually satisfied.\\

2) From 1), we can write $\|\x^t - \x^*\| \leq C{a}^t$ for some $a \in (0, 1)$ and some $C>0$. Considering $t \geq \bar{t}$, with $\bar{t}$ the first iteration for which both conditions (\ref{eq:condForEta1}) and (\ref{eq:condContraction}) are satisfied, we consider ${c}_t$ as in (\ref{eq:lineConvBound}), and let $T = t-k_t$\vspace{-0.2cm}
\begin{equation}\label{eq:boundB}\vspace{-0.2cm}
	\begin{aligned}
		{c}_t &\leq 1-\frac{\bar{\lambda}_{n,t}}{\bar{\rho}_t} + \frac{L}{\bar{\rho}_t}\|\boldsymbol{\theta}^{k_t} - \boldsymbol{\theta}^*\| + \frac{3L}{2\Bar{\rho}_t}\|\boldsymbol{\theta}^t - \boldsymbol{\theta}^*\|\\&\leq
		1-\frac{\bar{\lambda}_{n,t}}{\bar{\rho}_t} + \frac{L}{\bar{\rho}_t}C_1{a_*}^{t-T} + \frac{3L}{2\Bar{\rho}_t}C_2{a_*}^{t}\\&=
		1-\frac{\bar{\lambda}_{n,t}}{\bar{\rho}_t} + Ba_*^{t},
	\end{aligned}
\end{equation}
where $B = \frac{L}{\bar{\rho}_t}C_1a_*^{-T} + \frac{3L}{2\Bar{\rho}_t}C_2$, and $C_1, C_2$ are some bounded positive constants. For any iteration $t$, we can write $\|\x^{t+1} - \x^*\| \leq \bar{c}_t\|\x^t - \x^*\|$ for some $\bar{c}_t$ that could also be greater than one, if $t < \bar{t}$, but it is easy to see that $\bar{c}_t$ is always bounded. Now, we consider $\bar{a} := \limsup_{t}{(\prod_{k = 1}^{t}{\bar{c}_k})^{1/t}}$. It is straightforward to see that, as in the least squares case, it is $a_* \leq\bar{a}$. We can write $\log{\bar{a}} = \limsup_{t}{\frac{1}{t}}(\sum_{k = \bar{t} + 1}^{t}\log{{c}_k})$.
We get\vspace{-0.5cm}
\begin{equation*}\vspace{-0.2cm}
	\begin{aligned}
		\log{\bar{a}} &\leq \limsup_{t}{\frac{1}{t}}(\sum_{k = \bar{t} + 1}^{t}\log{{c}_k}) \\&\leq \limsup_{t}\frac{1}{t}\sum_{k = \bar{t} +1}^{t}\log{(1-\frac{\bar{\lambda}_{n,k}}{\bar{\rho}_k} + B{a_*}^k)}\\&= \limsup_{t}\frac{1}{t}\sum_{k = \bar{t} +1}^{t}\log{(1-\frac{\bar{\lambda}_{n,k}}{\bar{\rho}_k})} + \log{(1 + \frac{B{a_*}^{k}}{1 - \frac{\bar{\lambda}_{n,k}}{\bar{\rho}_k}})}.
	\end{aligned}
\end{equation*}
We see that the last term is\vspace{-0.2cm}
\begin{equation*}\vspace{-0.2cm}
	\log{(1 + \frac{B{a_*}^k}{1 - \frac{\bar{\lambda}_{n,k}}{\bar{\rho}_k}})} \leq \frac{B{a_*}^k}{1 - \frac{\bar{\lambda}_{n,k}}{\bar{\rho}_k}}\leq {\bar{B}{a_*}^k}
\end{equation*}
that comes from the identity $\log{(1+x)} \leq x$, and where $\bar{B} = \max_{k}\frac{1}{{1 - \frac{\bar{\lambda}_{n,k}}{\bar{\rho}_k}}}$, bounded because $|\mathcal{X}_t = 0|, \ \ \forall{t}$, and thus $\bar{\rho}_k > \bar{\lambda}_{n,k}, \forall k$. Now, we see that\vspace{-0.4cm} 
\begin{equation*}\vspace{-0.3cm}
	\limsup_{t}{\frac{1}{t}}\sum_{k = 1}^{t}{\bar{B}{a_*}^k} = \limsup_{t}\frac{1}{t}{\bar{B}}(\frac{1 - {a_*}^{t+1}}{1 - a_*} - 1) = 0.
\end{equation*}
Hence, we get, using also the finiteness of $\bar{t}$,\vspace{-0.2cm}
\begin{equation}\label{eq:logIdentity}\vspace{-0.2cm}
	\log{\bar{a}} = \limsup_{t}\frac{1}{t}\sum_{k = 1}^{t}\log{(1-\frac{\bar{\lambda}_{n,k}}{\bar{\rho}_k})}.
\end{equation}
Now we use local Lipschitz continuity (Assumption \ref{ass:localLCont}) to conclude the proof. Lipschitz continuity of $f^{(i)}(\boldsymbol{\theta})$ implies that, for any $k \in \{1,..., n\}$, $|\lambda_k^{(i)}(\x) - \lambda_k^{(i)}(\x^*)| \leq \bar{L}\|\x - \x^*\|$, which in turn implies $\lambda_k^{(i)}(\x) \geq \lambda_k^{(i)}(\x^*) - \bar{L}\|\x - \x^*\|$ and $\lambda_k^{(i)}(\x) \leq \lambda_k^{(i)}(\x^*) + \bar{L}\|\x - \x^*\|$. For a proof of this result, see also \cite{bisgard2020analysis}, page 116, Theorem 4.25. It follows that \vspace{-0.4cm}
\begin{equation*}\vspace{-0.3cm}
	\begin{aligned}
		\log{\bar{a}} &\leq \limsup_{t}\frac{1}{t}\sum_{k = 1}^{t}\log{(1 - \frac{\bar{\lambda}_n^{o} - \bar{L}\|\x^{k_t} - \x^*\|}{\bar{\rho}_k^{o} + \bar{L}\|\x^{k_t} - \x^*\|})}\\&\stackrel{(1)}{=} \limsup_{t}\frac{1}{t}\sum_{k = 1}^{t}\log{(1 - \frac{\bar{\lambda}_n^{o} }{\bar{\rho}_k^{o} + \bar{L}\|\x^{k_t} - \x^*\|})}\\& = \limsup_{t}\frac{1}{t}\sum_{k = 1}^{t}\log{(\frac{\bar{\rho}_k^{o} - \bar{\lambda}_n^{o} + \bar{L}\|\x^{k_t} - \x^*\|}{\bar{\rho}_k^{o} + \bar{L}\|\x^{k_t} - \x^*\|})}\\&\leq  \limsup_{t}\frac{1}{t}\sum_{k = 1}^{t}\log{(1 - \frac{ \bar{\lambda}_n^{o}}{{\bar{\rho}_k^{o}}} + \frac{\bar{L}\|\x^{k_t} - \x^*\|}{\bar{\rho}_k^{o}})}\\&\stackrel{(2)}{=} \limsup_{t}\frac{1}{t}\sum_{k = 1}^{t}\log{(1 - \frac{ \bar{\lambda}_n^{o}}{{\bar{\rho}_k^{o}}})}
	\end{aligned}
\end{equation*}
where equalities (1) and (2) follow from calculations equivalent to the ones used to obtain (\ref{eq:logIdentity}).\\
3) Consider $c_t$ as it was defined and bounded in (\ref{eq:boundB}), $C_3>0$ a constant such that $\log{c_k} \leq C_3, \ \forall k$, and such that $\log{B} \leq C_3$. Let $\bar{a}$ be defined as before. Let $a \in (0, 1)$. We have \vspace{-0.4cm}
\begin{equation}\vspace{-0.3cm}
	\begin{aligned}
		\log{\bar{a}} &= \limsup_{t}\frac{1}{t} \sum_{k = 1}^{t}\log{c_k}\\&\leq \limsup_{t}\frac{1}{t}\sum_{k = 1}^{t}\log{(1-\frac{\bar{\lambda}_{n, k}}{\bar{\rho}_k} + Ba^k)}\\&= \limsup_t\frac{1}{t}(\sum_{k\notin\mathcal{X}_t}^{}\log{c_k} + \sum_{k\in\mathcal{X}_t}^{}\log{Ba^k})\\& \leq 2C_3 + \limsup_t\frac{1}{t}\log{a}\sum_{k \in\mathcal{X}_t}^{}k\\& \leq 2C_3 + \limsup_{t}\frac{1}{t}\log aC_4|\mathcal{X}_t|^2 \\&= 2C_3 + \limsup_{t}\log aC_4{h(t)}^2 = -\infty
	\end{aligned}
\end{equation}
where $C_4 > 0$ is some positive constant and the last equality follows because $\log a < 0$ and $\lim_th(t) = \infty$. We see that $0 \leq a_* \leq \bar{a} \leq 0$, which implies $a_* = 0$..\vspace{-0.2cm}
\section*{Proof of Lemma \ref{Th:stepSizeConvex}}\label{proof:lemmaSS}
The following proof is similar to the proof for the beginning of the quadratically convergent phase in  centralized Newton method by \cite{ConvexOptimization}, page 490-491. We start with some definitions: at iteration $t$, let $f({\x}^t), \mathbf{g}_t = \nabla f({\x}^t), \mathbf{H}_t = \nabla^2 f({\x}^t)$ be the cost, the gradient and the Hessian, respectively, computed at ${\x}^t$. Let $\mathbf{H}_{k_t} = \nabla^2 f({\x}^{k_t})$ be the Hessian at ${\x}^{k_t}$, and $\hat{\mathbf{H}}_t$ the global Hessian approximation of Algorithm~\ref{alg:2}. Let the NT descent direction be $\mathbf{p}_t = \hat{\mathbf{H}}_t^{-1}\mathbf{g}_t$. Define
\begin{equation}\label{eq:L9defs}
	\begin{aligned}
		\bar{\sigma}_t^2 &:= \mathbf{p}_t^T\mathbf{g}_t =  \mathbf{g}_t^T\hat{\mathbf{H}}_t^{-1}\mathbf{g}_t,\\
		{\sigma}_t^2 &:= \mathbf{p}_t^T{\nabla^2f({\x}^t)}\mathbf{p}_t = \mathbf{p}_t^T{\mathbf{H}}_t\mathbf{p}_t,\\
		\Tilde{f}(\eta) &:= f({\x}^t - \eta\mathbf{p}_t), \ \ \Tilde{f}(0) = f({\x}^t),\\
		\Tilde{f}'(\eta) &:= \frac{\partial \Tilde{f}(\eta)}{\partial \eta} = -\nabla {f}({\x}^t - \eta\mathbf{p}_t)^T\mathbf{p}_t, \\
		\Tilde{f}''(\eta) &:= \frac{\partial^2 \Tilde{f}(\eta)}{\partial \eta^2} = \mathbf{p}_t^T\nabla^2 f({\x}^t - \eta \mathbf{p}_t)\mathbf{p}_t.
	\end{aligned}
\end{equation}
Note that
\begin{equation}\label{eq:tildeInZero}
	\begin{aligned}
		\Tilde{f}'(0) &= -\mathbf{g}_t^T\mathbf{p}_t = -\bar{\sigma}_t^2,\\
		\Tilde{f}''(0) &= \mathbf{p}_t^T\mathbf{H}_t\mathbf{p}_t = \sigma_t^2.
	\end{aligned}
\end{equation}
Note that $\hat{\mathbf{H}}_t \geq \bar{\kappa}, \ \ \forall{t}$. Thanks to $L$-Lipschitz continuity, it holds that $\|\nabla^2f({\x}^t - \eta \mathbf{p}_t) - \nabla^2f({\x}^t)\| \leq \eta L\|\mathbf{p}_t\|$ and we have that
\begin{equation}\label{eq:Bound1L9}
	\begin{aligned}
		|\Tilde{f}''(\eta) - \Tilde{f}''(0)| &= \mathbf{p}_t^T(\nabla^2f({\x}^t - \eta \mathbf{p}_t) - \nabla^2f({\x}^t))\mathbf{p}_t \\&\leq \eta L\|\mathbf{p}_t\|^3\leq \eta L \frac{\bar{\sigma}^3({\x}^t)}{\bar{\kappa}^{3/2}}
	\end{aligned}
\end{equation}
where the last inequality holds because $\bar{\kappa}\|\mathbf{p}_t\|^2 \leq \mathbf{p}_t^T\hat{\mathbf{H}}\mathbf{p}_t = \bar{\sigma}_t^2$. From (\ref{eq:Bound1L9}) it follows that 
\begin{equation}
	\Tilde{f}''(\eta) \leq \Tilde{f}''(0) + \eta L \frac{\bar{\sigma}^3({\x}^t)}{\bar{\kappa}^{3/2}} = \sigma_t^2 + \eta L \frac{\bar{\sigma}^3({\x}^t)}{\bar{\kappa}^{3/2}}.
\end{equation}
Similarly to \cite{ConvexOptimization}, page 490-491, we can now integrate both sides of the inequality getting 
\begin{equation*}
	\Tilde{f}'(\eta) \leq \Tilde{f}'(0)  + \eta\sigma_t^2 + \frac{\eta^2}{2}L\frac{\Bar{\sigma}_t^3}{\bar{\kappa}^{3/2}} = -\bar{\sigma}_t^2 + \eta\sigma_t^2 + \frac{\eta^2}{2}L\frac{\Bar{\sigma}_t^3}{\bar{\kappa}^{3/2}}.
\end{equation*}
By integrating again both sides of the inequality, we get, recalling that $\Tilde{f}(0) = f({\x}^t)$,
\begin{equation}\label{eq:Bound3L9}
	\begin{aligned}
		\Tilde{f}(\eta) &= f({\x}^t - \eta\mathbf{p}_t) \leq f({\x}^t) -\eta\Bar{\sigma}_t^2 + \frac{\eta^2}{2}\sigma_t^2 + \frac{\eta^3}{6}L\frac{\Bar{\sigma}_t^3}{\bar{\kappa}^{3/2}}.
	\end{aligned}
\end{equation}
Now, recalling $\mathbf{H}({\x}^{k_t}) \leq \hat{\mathbf{H}}_t$, we get that
\begin{equation}\label{eq:boundL9}
	\begin{aligned}
		\sigma_t^2 &= \mathbf{p}_t^T\mathbf{H}_t\mathbf{p}_t = \mathbf{p}_t^T(\mathbf{H}_{k_t} + \mathbf{H}_t - \mathbf{H}_{k_t})\mathbf{p}_t \\&=  \mathbf{p}_t^T\mathbf{H}_{k_t}\mathbf{p}_t + \mathbf{p}_t^T(\mathbf{H}_t - \mathbf{H}_{k_t})\mathbf{p}_t\\&\leq \mathbf{p}_t^T\hat{\mathbf{H}}_t\mathbf{p}_t + L\|\mathbf{p}_t\|^2\|{\x}^t - {\x}^{k_t}\| \leq \Bar{\sigma}_t^2 + \frac{L\Bar{\sigma}_t^2}{\bar{\kappa}}\|{\x}^t - {\x}^{k_t}\|,
	\end{aligned}
\end{equation}
where we have used Lipschitz continuity, and the fact that $\mathbf{p}_t^T\hat{\mathbf{H}}_t\mathbf{p}_t = \Bar{\sigma}_t^2$ and $\bar{\kappa}\|\mathbf{p}_t\|^2 \leq {\Bar{\sigma}_t^2}$. \newline
Next, setting $\eta = 1$ and plugging (\ref{eq:boundL9}) in (\ref{eq:Bound3L9}), we get 
\begin{equation}\label{eq:IneqArmijo}
	\begin{aligned}
		f({\x}^t - \mathbf{p}_t) &\leq f({\x}^t) -\Bar{\sigma}_t^2 + \frac{\sigma_t^2}{2} + \frac{L}{6}\frac{\Bar{\sigma}_t^3}{\bar{\kappa}^{3/2}}
		\\&\leq f({\x}^t) - \frac{\Bar{\sigma}_t^2}{2} + \frac{L\Bar{\sigma}_t^2}{2\bar{\kappa}}\|{\x}^t - {\x}^{k_t}\| + \frac{L}{6}\frac{\Bar{\sigma}_t^3}{\bar{\kappa}^{3/2}}
		\\&\leq f({\x}^t) - {\Bar{\sigma}_t^2}(\frac{1}{2} - \frac{L}{2\bar{\kappa}}\|{\x}^t - {\x}^{k_t}\| - \frac{L}{6}\frac{\Bar{\sigma}_t}{\bar{\kappa}^{3/2}})\\&=
		f({\x}^t) - \mathbf{p}_t^T\mathbf{g}_t(\frac{1}{2} - \frac{L}{2\bar{\kappa}}\|{\x}^t - {\x}^{k_t}\| - \frac{L}{6}\frac{\Bar{\sigma}_t}{\bar{\kappa}^{3/2}}).
	\end{aligned}
\end{equation}
In order for (\ref{eq:IneqArmijo}) to satisfy the Armijo-Goldstein condition (\ref{eq:Armijo}) for any parameter $\alpha \in (0, 1/2)$ we see that
\begin{equation}
	\frac{1}{2} - \frac{L}{2\bar{\kappa}}\|{\x}^t - {\x}^{k_t}\| - \frac{L}{6}\frac{\Bar{\sigma}_t}{\bar{\kappa}^{3/2}} \geq \alpha
\end{equation}
provides a sufficient condition. The above inequality can be written as
\begin{equation}
	3\bar{\kappa}\|{\x}^t - {\x}^{k_t}\| + \bar{\kappa}^{1/2}{\bar{\sigma}_t} \leq \frac{3\bar{\kappa}^2}{L}(1 - 2\alpha).
\end{equation}
We have that $\bar{\sigma}_t^2 = \mathbf{g}_t^T\hat{\mathbf{H}}^{-1}\mathbf{g}_t \leq \|\mathbf{g}_t\|^2/\bar{\kappa}$, which implies $\bar{\kappa}^{1/2}{\|}\sigma_t{\|} \leq \|\mathbf{g}_t\|$. Furthermore, by triangular inequality, we have $\|{\x}^t - {\x}^{k_t}\| \leq \|{\x}^t - {\x}^*\| + \|{\x}^{k_t} - {\x}^*\|$. Therefore, we see that if 
\begin{equation}\label{eq:conditionWeaker}
	3\bar{\kappa}(\|{\x}^t - {\x}^*\| + \|{\x}^{k_t} - {\x}^*\|) + \|\mathbf{g}_t\| \leq \frac{3\bar{\kappa}^2}{L}(1 - 2\alpha),
\end{equation}
then the Armijo-Goldstein condition is satisfied and $\eta = 1$ is chosen by the backtracking algorithm, proving the Lemma. Indeed, by $K$-smoothness of the cost function (see Assumption \ref{ass:localLCont}) we have $\|\mathbf{g}_t\| \leq K\|{\x}^t - {\x}^*\|$ and so the condition of the Lemma implies (\ref{eq:conditionWeaker}).

\newpage

%% file: JMLR/Appendix_B.tex
\section*{Results on EMNIST digits and w8a}
\begin{figure}[h!]
	\centering
	\includegraphics[width=0.85\columnwidth]{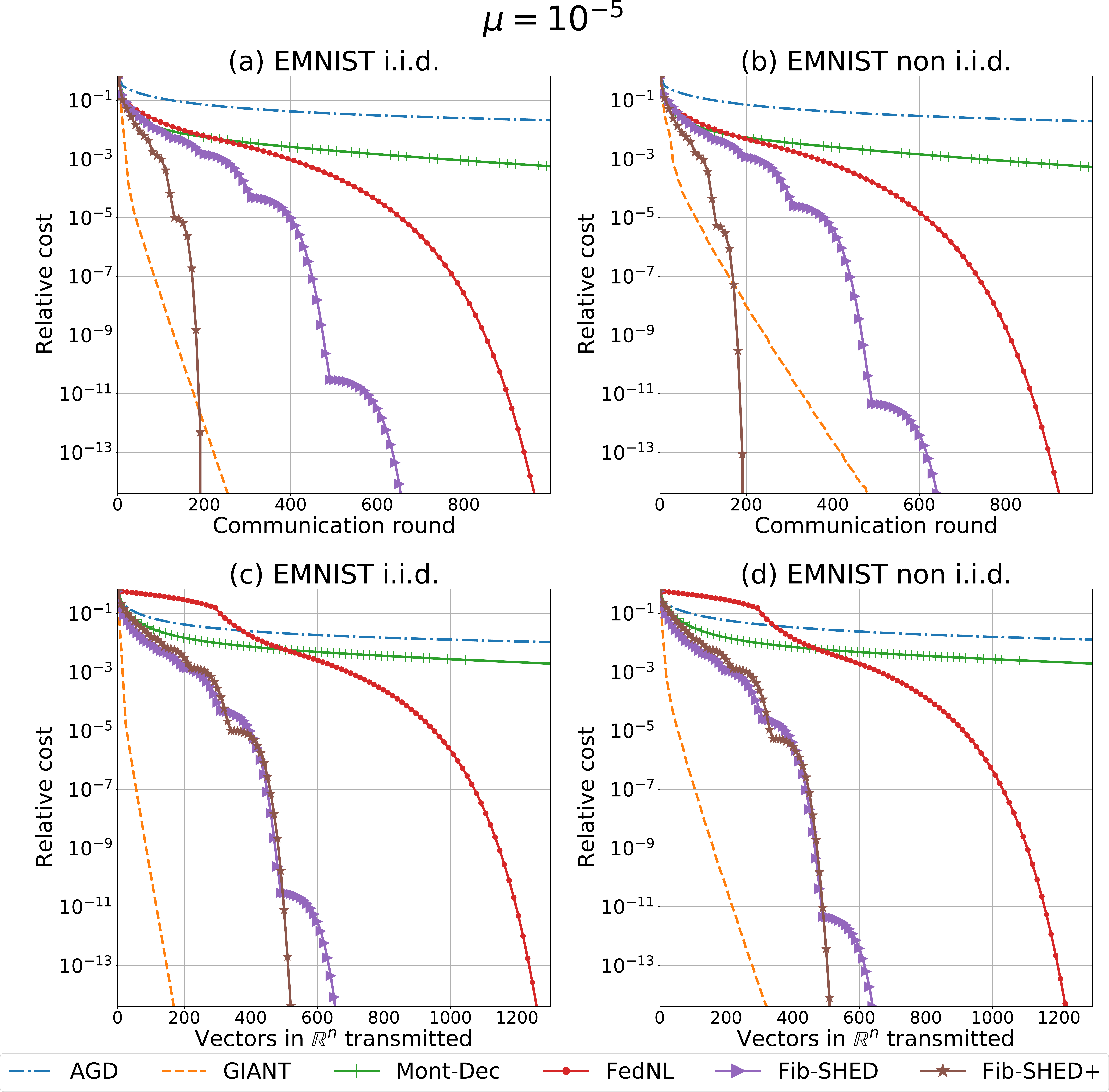}
	\caption{Performance comparison of logistic regression on EMNIST when $\mu = 10^{-5}$. Relative cost is $f(\theta^t) - f(\theta^*)$.}
	\label{fig:EMNIST_CompMu5}
\end{figure}

\begin{figure}[h!]
	\centering
	\includegraphics[width=0.85\columnwidth]{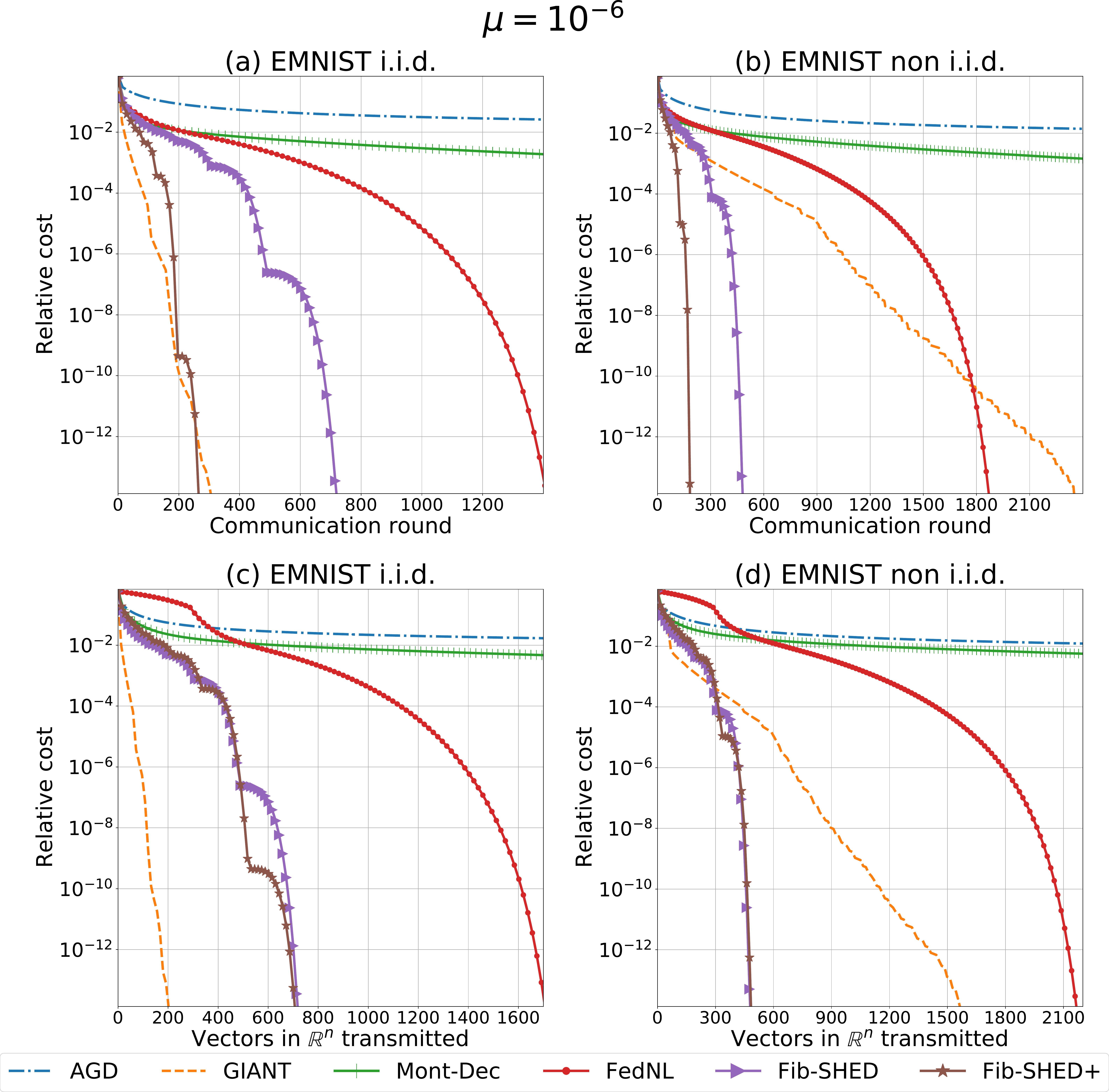}
	\caption{Performance comparison of logistic regression on EMNIST when $\mu = 10^{-6}$. Relative cost is $f(\theta^t) - f(\theta^*)$.}
	\label{fig:EMNIST_CompMu6}
\end{figure}

In this appendix, we include the results on the EMNIST digits and ‘w8a’ datasets when comparing the different algorithms. We show results for two values of the regularization parameter $\mu$, specifically $\mu = 10^{-5}$ and $\mu = 10^{-6}$, in Figure \ref{fig:EMNIST_CompMu5} and \ref{fig:EMNIST_CompMu6}, respectively. The results obtained on the EMNIST digits dataset confirm the results that were obtained with FMNIST, with the difference that in the case $\mu = 10^{-5}$, GIANT is not much impacted by the considered non i.i.d. configuration. The results obtained with the ‘w8a’ dataset show that, while GIANT performance is largely degraded because of the non i.i.d. configuration, also in this case Fib-SHED and Fib-SHED+ significantly outperform FedNL in both communication rounds and communication load required for convergence.

\begin{figure}
    \centering
    \includegraphics[width=0.85\columnwidth]{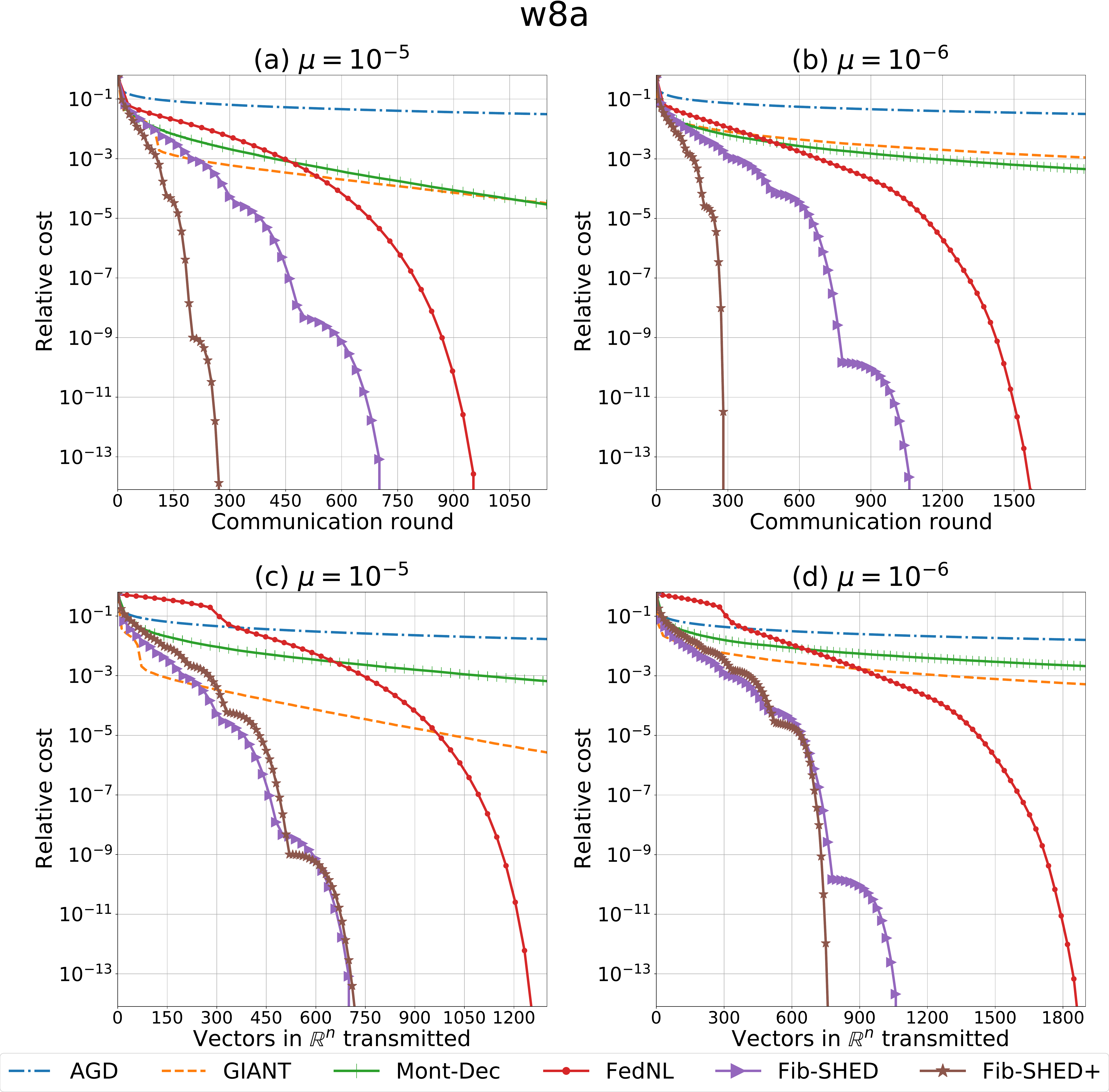}
    \caption{Performance comparison of logistic regression on w8a when $\mu = 10^{-5}$ and $\mu = 10^{-6}$. Relative cost is $f(\theta^t) - f(\theta^*)$.}
    \label{fig:w8a}
\end{figure}